%% file: tkde.tex
\documentclass[10pt,journal,compsoc]{IEEEtran}

\usepackage{graphicx}
\usepackage{amsmath}
\usepackage{amsthm}
\usepackage{booktabs}
\usepackage{amssymb}
\usepackage{array}
\usepackage{siunitx}
\usepackage{hyperref}
\usepackage{xcolor}
\usepackage{xspace}
\usepackage{longtable}
\usepackage{marvosym}
\usepackage{enumerate}
\usepackage{bm}
\usepackage{mathrsfs}
\usepackage{amsfonts}
\usepackage{multirow}
\usepackage[numbers]{natbib}
\usepackage{tikz-qtree}
\usepackage{pifont}
\usepackage{forest}
\usepackage{subcaption}

\newtheorem{definition}{Definition}

\usepackage{colortbl} 
\usepackage{xcolor} 
\definecolor{hidden-red}{RGB}{205, 44, 36}
\definecolor{hidden-blue}{RGB}{194,232,247}
\definecolor{hidden-orange}{RGB}{243,202,120}
\definecolor{hidden-green}{RGB}{34,139,34}
\definecolor{hidden-pink}{RGB}{255,245,247}
\definecolor{hidden-black}{RGB}{20,68,106}
\definecolor{LightRed}{rgb}{1,0.92,0.92}
\definecolor{LightOrange}{rgb}{1,0.95,0.88}
\definecolor{LightYellow}{rgb}{1.0,1.0,0.84}
\definecolor{LightGreen}{rgb}{0.9,1.0,0.88}
\definecolor{LightCyan}{rgb}{0.9,1,1}
\definecolor{LightBlue}{rgb}{0.9,0.94,1}
\definecolor{LightIndigo}{rgb}{0.92,0.9,1}
\definecolor{LightMagenta}{rgb}{0.96,0.86,1}
\definecolor{DirtyWhite}{rgb}{0.96,0.96,0.96}
\definecolor{DRed}{RGB}{255,0,0}
\usepackage{amssymb}
\usepackage{pifont}
\newcommand{\xmark}{\ding{55}}%

\newcommand{\etal}{\textit{et al}.}

\begin{document}

\title{Emerging Synergies in Causality and Deep Generative Models: A Survey}

\author{Guanglin~Zhou, 
    Shaoan~Xie,
    Guang-Yuan~Hao,
    Shiming~Chen,
    Biwei~Huang,
    Xiwei~Xu, 
    Chen~Wang, 
    Liming~Zhu, 
    Lina~Yao,
    \IEEEmembership{Senior~Member,~IEEE}, 
    Kun~Zhang
\IEEEcompsocitemizethanks{
\IEEEcompsocthanksitem G. Zhou is with the University of New South Wales.
Email: guanglin.zhou@unsw.edu.au
\IEEEcompsocthanksitem S. Xie is with Carnegie Mellon University. Email: shaoan@cmu.edu
\IEEEcompsocthanksitem G. Hao is with Mohamed bin Zayed University of Artificial Intelligence. Email: guangyuanhao@outlook.com
\IEEEcompsocthanksitem S. Chen is with Carnegie Mellon University and Mohamed bin Zayed University of Artificial Intelligence. Email: gchenshiming@gmail.com 
\IEEEcompsocthanksitem B. Huang is with University of California, San Diego. Email: bih007@ucsd.edu
\IEEEcompsocthanksitem X. Xu, C. Wang, L. Zhu and L. Yao are with CSIRO's Data61. Email: \{xiwei.xu, chen.wang, liming.zhu, lina.yao\}@data61.csiro.au
\IEEEcompsocthanksitem K. Zhang is with Carnegie Mellon University and Mohamed bin Zayed University of Artificial Intelligence. Email: kunz1@cmu.edu.
}}

\markboth{Journal of \LaTeX\ Class Files,~Vol.~14, No.~8, August~2021}%
{Shell \MakeLowercase{\textit{et al.}}: A Sample Article Using IEEEtran.cls for IEEE Journals}

\IEEEtitleabstractindextext{%
\begin{abstract}
In the field of artificial intelligence (AI), the quest to understand and model data-generating processes (DGPs) is of paramount importance.
Deep generative models (DGMs) have proven adept in capturing complex data distributions but often fall short in generalization and  interpretability.
On the other hand, causality offers a structured lens to comprehend the mechanisms driving data generation and highlights the causal-effect dynamics inherent in these processes.
While causality excels in interpretability and the ability to extrapolate, it grapples with intricacies of high-dimensional spaces.
Recognizing the synergistic potential, we delve into the confluence of causality and DGMs.
We elucidate the integration of causal principles within DGMs, investigate causal identification using DGMs, and navigate an emerging research frontier of causality in large-scale generative models, particularly generative large language models (LLMs).
We offer insights into methodologies, highlight open challenges, and suggest future directions, positioning our comprehensive review as an essential guide in this swiftly emerging and evolving area.
\end{abstract}
}

\maketitle
\IEEEdisplaynontitleabstractindextext
\IEEEpeerreviewmaketitle

\section{Introduction}
\IEEEPARstart{U}{nderstanding} and accurately modeling data-generating processes (DGPs) are pivotal objectives within the domains of artificial intelligence (AI) and machine learning (ML) \cite{pearl2009causal,Goodfellow2014GenerativeAN,Kingma2014AutoEncodingVB}.
The proficient modeling of these processes is not merely foundational but also has far-reaching implications across diverse applications.
It plays a crucial role in data analytics \cite{shalev2014understanding,kaur2022modeling,zhou2023contrastive}, synthesizing novel and high-fidelity data samples \cite{kobyzev2020normalizing,Dhariwal2021DiffusionMB,wang2022controllable}, as well as facilitating informed and reliable decision-making processes \cite{pearl2009causal,arora2021survey}.
As a result, various methodologies have emerged to tackle the multifaceted challenges inherent to this pivotal area of research \cite{Goodfellow-et-al-2016,Kingma2014AutoEncodingVB,pearl2009causal}.

\begin{figure}[!tb]
    \centering
    \includegraphics[width=0.99\linewidth]{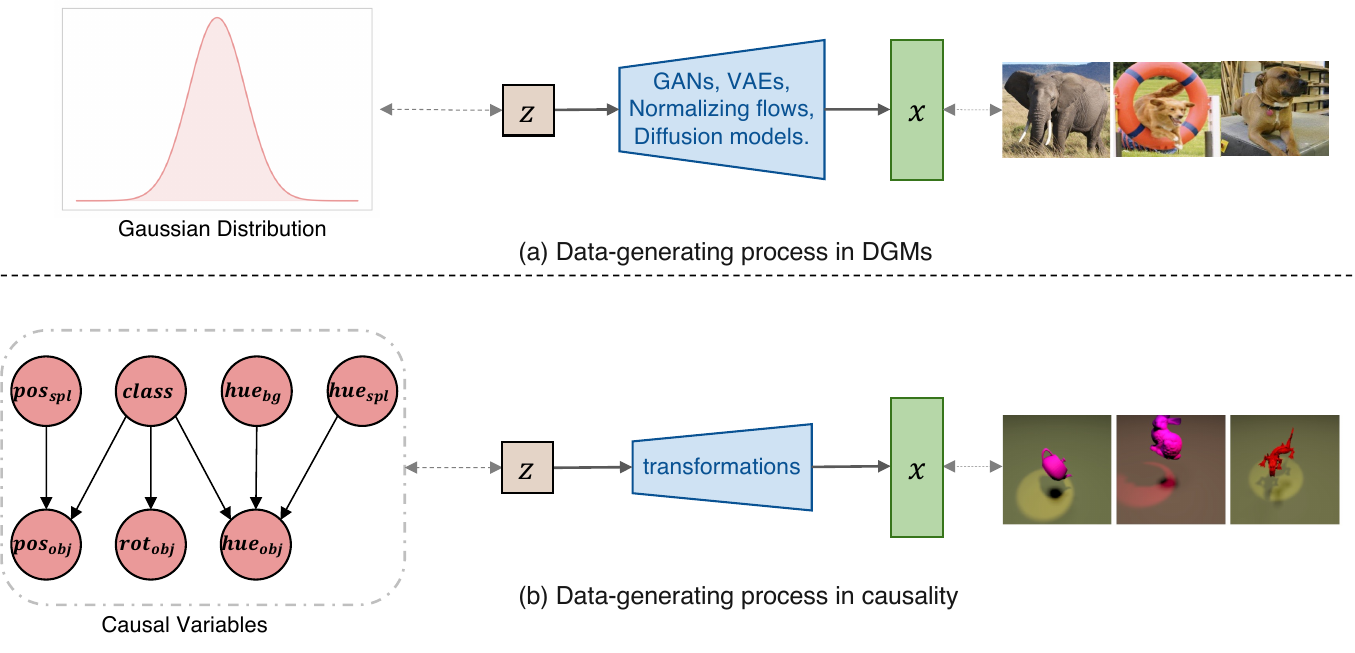} 
    \caption{
    Comparison of data-generating processes within deep generative models (DGMs) and causality. 
    DGMs primarily draw latent variables from a simple distribution such as Gaussian \cite{Goodfellow2014GenerativeAN, Kingma2014AutoEncodingVB}, while causality is rooted in variables defined by causal relationships \cite{von2021self}. 
    This divergence in foundation grants causality superior extrapolation and interpretability, and DGMs an edge in managing high-dimensional spaces, indicating synergistic opportunities.}
    \label{178756733687}
\end{figure}

Deep generative models (DGMs), encompassing architectures like generative adversarial networks (GANs) \cite{Goodfellow2014GenerativeAN,Mirza2014ConditionalGA,Chen2016InfoGANIR}, variational autoencoders (VAEs) \cite{Kingma2014AutoEncodingVB,higgins2016beta}, normalizing flows~\cite{dinh2016density,kingma2018glow,kobyzev2020normalizing}, and diffusion Models \cite{ho2020denoising,Song2021ScoreBasedGM,weng2021diffusion}, have proven to be indispensable in approximating intricate data distributions \cite{BondTaylor2022DeepGM}. 
GANs model DGPs using a game-theoretic approach to generate synthetic data nearly identical to real-world observations, while VAEs leverage a probabilistic framework to capture the data distribution, guided by latent variables.
Normalizing flows employ a series of invertible transformations to shape simple distributions into intricate ones, while diffusion models treat the DGPs as stochastic differential equations, evolving a basic noise distribution into a sophisticated data counterpart.
A prevalent pattern in DGMs is the transformation of a latent variable, typically sampled from a simple distribution like the Gaussian, into a complex sample that reflects the training data distribution, as illustrated in Figure \ref{178756733687}.
However, despite its pivotal role in generation, the latent variable often lacks clear or tangible interpretation, leading to challenges such as difficulties in out-of-sample extrapolation and inadequate learning of disentangled representations \cite{zhang2020causal,Kaddour2022CausalML}.

Unlike the abstract nature of latent variables in DGMs, causality offers a framework for understanding the underlying mechanisms that govern data-generating processes \cite{pearl1988probabilistic,pearl2009causal}. 
For example, structural causal models (SCMs) \cite{pearl2018book}, a cornerstone in causal theory, represent these mechanisms by defining causal relationships between variables, often formalized through directed acyclic graphs (DAGs), as shown in Figure \ref{178756733687}. 
This causal framework allows for meaningful interventions and counterfactual analyses, offering a level of rigor often missing in DGMs. 
Nevertheless, causality is not without its limitations.
Traditional causal models may find it challenging to deal with high-dimensional unstructured data that DGMs handle with relative ease. 
Identifying causal structures in such settings can be computationally intensive and may result in identifiability issues \cite{Hyvrinen1999NonlinearIC, Glymour2019ReviewOC}.
Herein lies the opportunity for synergy: DGMs can offer powerful computational tool to approximate these complex and high-dimensional data, thereby complementing the strengths of causal models.

Recognizing the unique yet complementary capabilities of DGMs and causality in shaping DGPs, our survey examines their collaborative potential: integrating causal principles in DGMs and identifying causality via DGMs.
Furthermore, we delve into an emerging overlap of causality with large-scale generative models \cite{oppenlaender2022creativity,ramesh2022hierarchical,rombach2022high}, specifically generative large language models (LLMs) \cite{brown2020language,ouyang2022training,radford2018improving,radford2019language}.
We posit that the fusion might lead to the develpment of more robust and interpretable generative AI systems \cite{cao2023reinforcement,jo2023promise}.

\textbf{Motivations of this survey.}
While extensive literatures exists on causality \cite{pearl2009causal,yao2021survey,Kaddour2022CausalML} and DGMs \cite{oussidi2018deep,harshvardhan2020comprehensive,bond2021deep} independently, they typically address distinct facets of their respective domains.
Causality-based surveys primarily concentrate on foundational theories \cite{pearl2009causality} and methodologies for reasoning and inference \cite{yao2021survey}.
Though one survey touch upon the overlap of causality with machine learning \cite{Kaddour2022CausalML}, such work rarely provides an exhaustive exploration with DGMs.
On the other hand, DGMs-centric surveys typically highlight model structures and training strategies, frequently overlooking the integral role of causality \cite{oussidi2018deep,harshvardhan2020comprehensive,bond2021deep}.
This scenario underscores a notable gap: the absence of a comprehensive study that explores the confluence of causality and DGMs.
Bridging this gap holds potential for advancements like generative models with enhanced interpretability and superior generalization.
Recognizing this potential, our survey endeavors to offer a comprehensive exploration of this emerging and promising intersection.

\begin{figure}[!tb]
	\centering
	\includegraphics[scale=0.43]{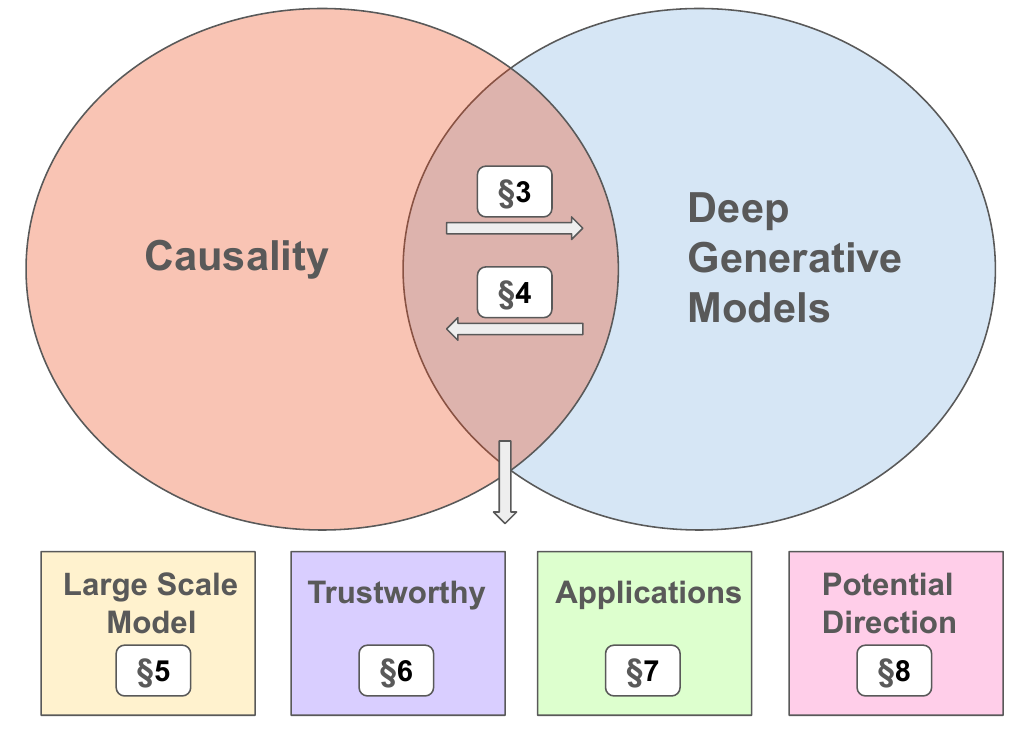}
	\caption{The illustration of the intersections of causality and deep generative models (DGMs). 
 \S \ref{971699495725} examines causal principles in DGMs, \S \ref{776815666820} explores DGMs in causal tasks, \S \ref{605156580006} discusses causality in large-scale DGMs, \S \ref{471209836872} addresses enhancing DGM trustworthiness through causality, \S \ref{591853874791} covers varied applications of causality with DGMs, and \S \ref{996252365804} provides potential directions.}
	\label{684881581452}
\end{figure}

\textbf{Survey organizations and contributions.}
Our contributions are threefold.   
Firstly, to the best of our knowledge, we provide the first exhaustive review of the synergies between causality and DGMs.
Specifically, we delve into integrating causal principles into DGMs (\S \ref{971699495725}) and further explore the potential of identifying causality via DGMs (\S \ref{776815666820}).
Secondly, we shed light on the nascent research area involving causality and large-scale generative models, with a special emphasis on generative large language models (\S \ref{605156580006}). 
Our goal is to highlight the potential pathways for crafting more robust and interpretable generative AI. 
Lastly, we provide an overview in  trustworthy properties (\S \ref{471209836872}) and applications (\S \ref{591853874791}), and provide open challenges and prospective directions for future research (\S \ref{996252365804}).
The overlapping aspects of causality and DGMs is depicted in Figure \ref{684881581452}.

\section{Background}
\label{241725040005}
\input{secs/sec_Backgrond}

\begin{table*}[!th]
\caption{A detailed categorization of the synergy between causality and DGMs, considering generative architectures, application tasks, and trustworthiness properties. The table also indicates the datasets used for testing the reviewed models.}
\setlength\tabcolsep{2pt}
\centering
\begin{tabular}{>{\centering\arraybackslash}p{3.6cm}|>{\centering\arraybackslash}p{2.5cm}|>{\centering\arraybackslash}p{3.9cm}|>{\centering\arraybackslash}p{2.2cm}|>{\centering\arraybackslash}p{4.5cm}}
\toprule
\textbf{Paper} & \textbf{Architecture} & \textbf{Tasks} & \textbf{Trustworthy Properties} & \textbf{Data sets}  \\
\midrule
\rowcolor{LightRed}
\textbf{Integrating Causal Principles in DGMs} &&&&\\ \hline
\rowcolor{LightRed}
CausalGAN~\cite{Kocaoglu2018CausalGANLC} & GANs & image generation & generalization & CelebA~\cite{liu2015faceattributes} \\ \hline
\rowcolor{LightRed}
CGN~\cite{Sauer2021CounterfactualGN} & GANs & multi-task (image generation, classification) & generalization &  MNISTs~\cite{Kim2019LearningNT,cimpoi2014describing}, ImageNet~\cite{imagenet_cvpr09}  \\ \hline
\rowcolor{LightRed}
CausalTGAN~\cite{Wen2021CausalTGANGT} & GANs & tabular data generation & generalization & Adult\footnotemark[1], Census\footnotemark[2], Cabs\footnotemark[3], Loan\footnotemark[4], News\footnotemark[5], Kings\footnotemark[6]\\  \hline
\rowcolor{LightRed}
CFGAN~\cite{Xu2019AchievingCF} & GANs &tabular data generation & fairness & Adult\footnotemark[1] \\ \hline
\rowcolor{LightRed}
DECAF~\cite{Breugel2021DECAFGF} & GANs & tabular data generation & fairness & Census\footnotemark[2], Credit \cite{Asuncion2007UCIML} \\ \hline
\rowcolor{LightRed}
DEAR~\cite{shen2022weakly} & GANs & multi-task (image generation, classification) & generalization, interpretability & Pendulum~\cite{Yang2021CausalVAEDR}, CelebA~\cite{liu2015faceattributes}  \\ \hline
\rowcolor{LightRed}
GenInt~\cite{Mao2021GenerativeIF} & GANs & multi-task (image generation, classification) & generalization & ImageNet~\cite{imagenet_cvpr09}, ObjectNet~\cite{Barbu2019ObjectNetAL}, ImageNet-C~\cite{hendrycks2019benchmarking}, ImageNet-V2~\cite{recht2019imagenet}\\ \hline
\rowcolor{LightRed}
 iStyleGAN~\cite{xie2022multi}  & GANs & multi-task (causal representation learning, image generation and translation) & generalization  & CelebA~\cite{liu2015faceattributes}, AFHQ~\cite{choi2020stargan} \\
 \rowcolor{LightRed}
iMSDA~\cite{kong2022partial} & VAEs, Normalizing Flows & multi-task (causal representation learning, domain adaptation) & generalization & PACS~\cite{li2017deeper}, OfficeHome~\cite{venkateswara2017deep}\\ \hline
\rowcolor{LightRed}
PKD~\cite{Feng2022PrincipledKE} & GANs & image generation & generalization & ImageNet~\cite{imagenet_cvpr09}, FFHQ~\cite{Karras2019ASG}\\ \hline
\rowcolor{LightRed}
CausalVAE~\cite{Yang2021CausalVAEDR} & VAEs & image generation & generalization, interpretability& CelebA~\cite{liu2015faceattributes} \\ \hline
\rowcolor{LightRed}
CounterfactualMS\cite{Reinhold2021ASC} & VAEs & image generation & interpretability & -- \\ \hline

\rowcolor{LightRed}
Causal-gen~\cite{ribeiro2023high} & VAEs & image generation & interpretability & Morpho-MNIST~\cite{castro2019morpho}, brain MRI scans~\cite{Sudlow2015UKBA}, Chest X-ray~\cite{johnson2019mimic} \\ \hline
\rowcolor{LightRed}
Hu \etal~\cite{Hu2021ACL} & VAEs, GPT\footnotemark[8] & text generation & fairness, interpretability & YELP\footnotemark[7], BIOS corpus~\cite{de2019bias} \\ \hline
\rowcolor{LightRed}
Diff-SCM~\cite{Sanchez2022DiffusionCM} & Diffusion Models & image generation & generalization & MNIST~\cite{lecun1998gradient}, ImageNet~\cite{imagenet_cvpr09}  \\ \hline
\rowcolor{LightRed}
CDPM~\cite{Sanchez2022WhatIH} & Diffusion Models & multi-task (image generation, classification) & generalization, interpretability & BraTS~\cite{bakas2018identifying}\\ \midrule

\rowcolor{LightOrange}
\textbf{Identifying Causality via DGMs} &&&&  \\ \hline
\rowcolor{LightOrange}
GCIT~\cite{bellot2019conditional} & GANs & causal discovery & -- & cancer genome~\cite{garnett2012systematic} \\ \hline
\rowcolor{LightOrange}
SAM~\cite{kalainathan2022structural}& GANs &causal discovery & -- & Sachs~\cite{sachs2005causal}\\ \hline
\rowcolor{LightOrange}
GANITE~\cite{yoon2018ganite} & GANs & counterfactual inference & -- & IHDP~\cite{shalit2017estimating}, Jobs~\cite{smith2005does}, Twins~\cite{hannart2016causal}\\ \hline
\rowcolor{LightOrange}
SCIGAN~\cite{bica2020estimating} & GANs & counterfactual inference & -- & TCGA~\cite{weinstein2013cancer}, News~\cite{schwab2020learning}, MIMIC III~\cite{johnson2016mimic} \\ \hline
\rowcolor{LightOrange}
DeepSCM~\cite{Pawlowski2020DeepSC} & VAEs & multi-task (counterfactual inference, image generation) & generalization, interpretability & Morpho-MNIST~\cite{castro2019morpho}, brain MRI scans~\cite{Sudlow2015UKBA} \\ \hline
\rowcolor{LightOrange}
VACA~\cite{SnchezMartn2021VACADO} & VAEs & counterfactual inference & fairness &  Adult\footnotemark[1], Loan\footnotemark[4], Credit~\cite{Asuncion2007UCIML} \\ \hline
\rowcolor{LightOrange}
CAREFL~\cite{khemakhem2021causal} & Normalizing Flows & causal discovery & -- & Tubingen cause-effect~\cite{mooij2016distinguishing}, EEG~\cite{dornhege2004boosting} \\ \hline
\rowcolor{LightOrange}
OCDaf~\cite{kamkari2023ocdaf} & Normalizing Flows & causal discovery & -- & Sachs~\cite{sachs2005causal} \\ \hline
\rowcolor{LightOrange}
BGM~\cite{nasr2023counterfactual} & Normalizing Flows & counterfactual inference & -- & --\\ \hline
\rowcolor{LightOrange}
DiffAN~\cite{sanchez2022diffusion} & Diffusion Models & causal discovery & -- & -- \\ \hline
\rowcolor{LightOrange}
DCM~\cite{chao2023interventional}  & Diffusion Models & counterfactual inference & -- & fMRI~\cite{thompson2020data} 
 \\ \midrule

\rowcolor{LightBlue}
\textbf{Exploring Causality within Large-scale DGMs} &&&&  \\ \hline
\rowcolor{LightBlue}
Hobbhahn \etal~\cite{hobbhahn2022investigating} & GPT\footnotemark[8] & event causality identification & interpretability & -- \\ \hline
\rowcolor{LightBlue}
Zhang \etal~\cite{zhang2023understanding} & GPT\footnotemark[8] & event causality identification& interpretability & -- \\ \hline
\rowcolor{LightBlue}
Nick \etal~\cite{pawlowski2023answering} & GPT\footnotemark[8] & event causality identification& interpretability & -- \\ \hline
\rowcolor{LightBlue}
K{\i}c{\i}man \etal~\cite{kiciman2023causal} & GPT\footnotemark[8] & causal discovery, event causality identification & interpretability & Tubingen cause-effect~\cite{mooij2016distinguishing}, Neuropathic pain~\cite{tu2019neuropathic} \\ \hline
\rowcolor{LightBlue}
Gao \etal~\cite{gao2023chatgpt}& GPT\footnotemark[8] & causal discovery, event causality identification, causal explanation & interpretability & e-CARE~\cite{du2022care}, COPA~\cite{roemmele2011choice}, EventStoryLine~\cite{caselli2017event}, Causal-TimeBank~\cite{mirza2014annotating}, MAVEN-ERE~\cite{wang2022maven} \\ \hline 
\rowcolor{LightBlue}
LMPriors~\cite{choi2022lmpriors} & GPT\footnotemark[8] & causal discovery & interpretability & Tubingen cause-effect~\cite{mooij2016distinguishing} \\ \hline
\rowcolor{LightBlue}
Long \etal~\cite{long2023can} & GPT\footnotemark[8] & causal discovery & interpretability & -- \\ \hline
\rowcolor{LightBlue}
Ban \etal~\cite{ban2023query} & GPT\footnotemark[8] & causal discovery & interpretability & -- \\ \hline
\rowcolor{LightBlue}
Long \etal~\cite{long2023causal} & GPT\footnotemark[8] & causal discovery & interpretability & Asia~\cite{lauritzen1988local}, CHILD~\cite{shafer1996probabilistic}, Insurance~\cite{binder1997adaptive} \\ \hline
\rowcolor{LightBlue}
Matej \etal~\cite{willig2023causal} & GPT\footnotemark[8] & causal discovery & interpretability & Tubingen cause-effect~\cite{mooij2016distinguishing} \\ \hline
\rowcolor{LightBlue}
Jin \etal~\cite{jin2023can} & GPT\footnotemark[8], LLaMa~\cite{touvron2023llama} & causal discovery & interpretability & corr2cause\footnotemark[9] \\ \hline
\rowcolor{LightBlue}
Jin \etal~\cite{jin2023cladder} & GPT\footnotemark[8], LLaMa~\cite{touvron2023llama} & causal inference & interpretability & cladder\footnotemark[10] \\ \bottomrule
\end{tabular}
\label{302764778028}
\end{table*}

\section{Integrating Causal Principles in DGMs}
\label{971699495725}
\input{secs/sec_Causality4DGMs}

\section{Identifying Causality via DGMs}
\label{776815666820}
\input{secs/sec_DGMs4Causality}

\section{Exploring Causality within Large-scale DGMs: An Emerging Frontier}
\label{605156580006}

\input{secs/sec_Causality_LLMs}

\section{Enhancing Trustworthy Properties in DGMs through Causality}
\label{471209836872}
{
The integration of causality in DGMs represents a pivotal advancement in enhancing their trustworthy properties, including generalization, fairness, and interpretability. 
According to \cite{Kaur2022TrustworthyAI}, these characteristics are essential for making machine learning models more reliable and ethically sound. 
In this section, we delve into how causal principles contribute to each of these trustworthy properties in DGMs.
}


\subsection{Generalization}
\label{552901325861}
Deep neural networks often suffer from challenges like shortcut learning and spurious correlations, which hinder their generalizability~\cite{Geirhos2020ShortcutLI}.
Causality, by rigorously modeling the data-generating process, offers a promising perspective for mitigating such issues \cite{castro2020causality,degrave2021ai}. 
The concept of \textbf{\textit{counterfactual}}, a key element in causality, involves modifying certain data features while preserving others constant, thus offering insights into alternative scenarios. 
In generative modeling, this approach enables the creation of models that are more resilient to environmental changes and less prone to spurious correlations, i.e., style and content decomposition~\cite{von2021self}, thereby enhancing generalization across diverse domains.
{
Enhancements in DGMs through causality are evident in various studies. \cite{Sauer2021CounterfactualGN} shows improved classifiers using causally-informed GAN samples. \cite{Mao2021GenerativeIF} and \cite{Zhang2021LearningCR} highlight robust data augmentation with causal semantics. \cite{shen2022weakly} demonstrates the benefits in distributional robustness. In medical imaging, \cite{Pawlowski2020DeepSC} and \cite{Sanchez2022WhatIH} utilize counterfactuals for improved brain MRI analysis, enhancing model interpretability and reliability.
}
\subsection{Fairness}
\label{689390515520}
{ The integration of causality into DGMs plays a crucial role in addressing fairness, a key trustworthy property.}
Machine learning models, can inadvertently amplify biases inherent in training data, leading to discriminatory decisions in critical applications such as insurance, hiring and lending.
Particularly, subgroups characterized by sensitive attributes like race and gender may face unjust biases \cite{kusner2017counterfactual}.
{ Causal fairness emerges as a solution, offering a framework to identify and mitigate these biases.}
Several studies, including CFGAN and DECAF have successfully enforced various causal fairness principles, building their methodologies upon foundational causal graph assumptions~\cite{Wu2019CounterfactualFU,Xu2019AchievingCF}.
{ Concurrently, approaches like VGAE \cite{SnchezMartn2021VACADO} have demonstrated the effectiveness of incorporating pre-defined causal structures in assessing counterfactual fairness and training unbiased classifiers.} 
{ Overall, these examples highlights the potential of causality in enhancing the fairness of machine learning models, underscoring its importance in responsible and unbiased decision-making.}

\subsection{Interpretability}
\label{109024743992}
{ DGMs aim to capture and manipulate underlying data variations. 
Traditional DGMs often rely on the assumption of statistical independence among latent variables, a notion that has been challenged in recent research \cite{Locatello2019ChallengingCA,Besserve2020CounterfactualsUT,xie2020generalized}. Incorporating causality into DGMs addresses this by establishing causal relationships between factors, leading to models that not only capture data variation but also reflect the underlying causal structure.
An illustrative example is the Pendulum dataset \cite{Yang2021CausalVAEDR}, where altering the light source position affects variables like shadow position and length, demonstrating inherent causal mechanisms. This use of a causal perspective enhances the interpretability of DGMs, allowing for a deeper understanding of how changes in one variable can influence others \cite{Pawlowski2020DeepSC}.
Furthermore, the integration of counterfactual explanations with DGMs provides a robust framework for comprehending classifier decisions. Studies like \cite{Augustin2022DiffusionVC,Khorram2022CycleConsistentCB,Jeanneret2022DiffusionMF,boreiko2022sparse} showcase how counterfactual reasoning can unravel the decision-making process of models, making them more transparent and interpretable.}

\section{Applications}
\label{591853874791}
{
This section explores the multifaceted applications of causality in conjunction with Deep Generative Models (DGMs) across various domains. By examining specific instances in healthcare, social sciences, and emerging interdisciplinary fields, we highlight the breadth and impact of this technological integration.}

\subsection{Healthcare}
\label{766991632840}
{ In healthcare, the integration of causality with DGMs has led to significant advancements.} 
CounteRGAN \cite{nemirovsky2020countergan}, showcases an enhancement in classifier accuracy when applied to tabular disease datasets, notably the Pima Indians Diabetes dataset\footnotemark[12] \cite{Smith1988UsingTA}. 
{ Neuroimaging also benefits from this technology, with DeepSCM \cite{Pawlowski2020DeepSC} applying interventions on factors like age, leading to the generation of credible counterfactual brain images.} 
Further, \cite{rasal2022deep} and \cite{Sanchez2022WhatIH} showcase the application of causality and DGMs in 3D brain structure analysis and medical image segmentation, respectively.

\subsection{Social Science}
\label{133894477284}
{ In the realm of social sciences, the application of causality in deep generative models has enabled significant progress, particularly in addressing biases.}
Xu \etal~\cite{Xu2019AchievingCF} introduce a fairness-aware GAN tailored for the UCI Adult income dataset \cite{Dua:2019}, aiming to neutralize gender biases. 
Similarly, Daniel \etal~\cite{nemirovsky2020countergan} use counterfactuals to tackle racial and gender biases in the COMPAS dataset \cite{compas-analysis}. 
Wen et al.~\cite{Wen2021CausalTGANGT} extended these concepts to a broader range of datasets. Additionally, \cite{Breugel2021DECAFGF} and \cite{Hu2021ACL} have made strides in fair credit decision-making and reducing sentiment biases within the YELP\footnotemark[7] dataset, respectively.
{ While significant strides have been made in addressing gender and racial biases in social sciences through causality in DGMs, research in other forms of bias such as ethnicity, age, religion, and language remains sparse. 
We acknowledge this as a notable gap in the literature and advocate for future research in these areas.}

\footnotetext[1]{\url{http://archive.ics.uci.edu/ml/datasets/adult}}
\footnotetext[2]{\url{https://archive.ics.uci.edu/ml/datasets/census+income}}
\footnotetext[3]{\url{https://tinyurl.com/6jnzyx9e}}
\footnotetext[4]{\url{https://tinyurl.com/2kf238vf}}
\footnotetext[5]{\url{https://archive.ics.uci.edu/ml/datasets/online+news+popularity}}
\footnotetext[6]{\url{https://www.kaggle.com/harlfoxem/housesalesprediction}}
\footnotetext[7]{\url{https://www.yelp.com/dataset/challenge}}
\footnotetext[8]{\url{https://platform.openai.com}}
\footnotetext[9]{\url{https://huggingface.co/datasets/causalnlp/corr2cause}}
\footnotetext[10]{\url{https://huggingface.co/datasets/tasksource/cladder}}
\footnotetext[11]{\url{https://github.com/rr-learning/disentanglement\_dataset}}
\footnotetext[12]{\url{https://www.kaggle.com/datasets/uciml/pima-indians-diabetes-database}}
\footnotetext[13]{\url{https://www.kaggle.com/datasets/harlfoxem/housesalesprediction}}
\footnotetext[14]{https://cdas.cancer.gov/datasets/plco/22/}

\subsection{Emerging and Interdisciplinary Applications}
{ Beyond healthcare and social sciences, DGMs with causality hold potential in fields such as IoT~\cite{hassan2019current}, smart cities~\cite{sanchez2019smart}, education~\cite{chen2020artificial}, and text recognition~\cite{chen2021text}. 
Particularly in language generation, advancements in LLMs demonstrate the impactful fusion of DGMs with causality, contributing to significant progress in natural language processing~\cite{openai2023gpt4, kiciman2023causal,hobbhahn2022investigating,jin2023cladder}. 
Incorporating causal understanding into LLMs can further enhance the logical coherence and contextual relevance of their outputs, leading to more sophisticated and accurate language models.
While current research in these areas is not as extensive, they represent important avenues for future exploration. 
}

\section{Discussion}
\label{996252365804}
In our survey, we have systematically explored the synergies between causality and Deep Generative Models (DGMs), shedding light on their complementary strengths and challenges in modeling Data-Generating Processes (DGPs). 
In this section, we highlight some limitations of current methodologies and suggest prospective research directions. 

\subsection{Understanding generated counterfactuals}
\label{528767552000}
Within generative modeling, counterfactual samples are conceived by modifying certain features or variables of the original data while preserving others, thus simulating what might have been under alternative conditions. 
A salient example in this context is the generation of images by maintaining the content but altering the style \cite{von2021self}. 
A central challenge in the realm of counterfactuals is their inherent unverifiability, making it intricate to ascertain the validity of generated counterfactual samples \cite{Pawlowski2020DeepSC,Reinhold2021ASC}. 
Pioneering efforts in this domain include CausalGAN \cite{Kocaoglu2018CausalGANLC}, which underscores the potential of causality to bolster the creative prowess of deep generative models. Specifically, it enables them to generate samples that deviate from training distributions, i.e., $\{Gender=Female, Mustache=1\}$. 
While CausalGAN showcases its capability to discern between conditional and counterfactual images on the CelebA dataset, it also underscores the lack of robust, practical metrics to evaluate generated counterfactuals.

Several studies have approached this challenge by gauging the utility of counterfactuals through proxy tasks. These tasks evaluate whether counterfactual samples enhance performance in downstream applications. For instance, some studies leverage counterfactuals to train invariant classifiers, bolster out-of-distribution robustness \cite{Sauer2021CounterfactualGN}, eliminate undesired spurious features, and enhance model resilience \cite{Mao2021GenerativeIF,Zhang2021LearningCR}. These efforts can be encapsulated under the umbrella of Counterfactual Data Augmentation (CFDA). However, the optimal conditions and extent for deploying CFDA remain largely unexplored \cite{wiles2021fine,Kaddour2022CausalML}.

Moving forward, the establishment of quantifiable metrics to evaluate generated counterfactuals is imperative \cite{Reddy2022OnCD}. Particularly in sensitive domains like healthcare, these evaluations might necessitate domain expertise and supervisory signals \cite{Reinhold2021ASC,Sanchez2022WhatIH}. The introduction of benchmark datasets, tailored for evaluating counterfactual reasoning methods, can accelerate advancements in this space. In line with this, recent efforts by Monteiro et al. \cite{monteiro2023measuring} delve into the axiomatic evaluation of counterfactual models, presenting a promising avenue for assessing image counterfactuals.

\subsection{Challenges with Large-scale Causal Graphs}
\label{413845799289}
A key limitation inherent to the current generation within CGMs is the capacity to handle scalability, especially when it comes to large-scale causal graphs. 
While many of the works reviewed in this survey, including those focused on image generation, have made commendable progress, they are typically constrained to causal structures with relatively few variables. 
As an illustration, CausalGAN \cite{Kocaoglu2018CausalGANLC}, which arguably presents one of the more intricate causal structures among the surveyed literature, still encapsulates a causal graph of no more than ten variables, each with a maximum degree of three, as shown in Figure \ref{040922319451}.

While certain studies, such as \cite{Sanchez2022DiffusionCM}, have transparently recognized and acknowledged the simplifications made in their SCMs, often restricting them to bi-variable SCMs, real-world datasets frequently present a far more convoluted landscape. 
For instance, the Colorectal Cancer Dataset\footnotemark[14] comprises a staggering 400k variables \cite{javidian2021scalable}. 
Consequently, scaling CGMs to accommodate these large-scale causal graphs remains a pivotal and pressing challenge warranting further exploration in upcoming research.

\subsection{Incorporating Causal Discovery for Causal Generative Models} 
\label{161379479878}

A salient requirement for most causal generative models is the prior knowledge of the underlying causal structure. 
Although some recent works have attempted to mitigate this constraint, they still necessitate the specification of causal variable labels and rely on supervised information for identifiability \cite{Yang2021CausalVAEDR,shen2022weakly,Kaddour2022CausalML}. 
Such requirements can pose challenges, especially in real-world situations where causal orderings might not be readily available. 
This highlights a promising avenue for future research: the integration of causal discovery methods \cite{huang2020causal,xie2020generalized,xie2022identification,huang2022latent}. These methods aim to discern causal variables and graph structures under structure identifiability constraints, offering a more autonomous approach to understanding causal relations \cite{shen2022weakly}.


\subsection{Causal Learning within Large-scale Generative Models}
\label{993776787668}
The integration of causality with large-scale generative models, especially generative LLMs, brings forth novel research avenues. 
As we delve into this confluence, three key considerations stand out:
\textbf{1) Need for a Unified Evaluation Framework:} 
The causal reasoning performance of LLMs varies based on the datasets employed, architectural nuances, and assessment metrics \cite{kiciman2023causal,gao2023chatgpt,jin2023can}. 
Establishing a standardized testbed \cite{gulrajani2020search,fu2023mme} becomes imperative to gauge these capabilities uniformly, offering a clearer perspective on where LLMs excel or fall short in causal contexts.
\textbf{2) Prioritizing Concept Learning:} 
Effective causal reasoning hinges on the deep-rooted understanding of concepts (an example of four concepts shown in Figure \ref{019405822627}) \cite{tenenbaum1998bayesian,lake2015human}. 
Fortifying LLMs' proficiency in discerning and connecting these fundamental ideas is vital. 
This endeavor is not just about refining current capabilities but seeks to redefine the paradigm of causal comprehension in generative AI.
\textbf{3) Moving Beyond Prompts:} 
While adeptly designed prompts can steer LLMs towards desired outputs, an over-reliance on them can mask inherent challenges \cite{long2023can,jin2023cladder}. 
It's essential to recognize and address the foundational issues within LLMs, like ChatGPT, encounter in causal reasoning, rather than merely navigating around them with prompts \cite{gao2023chatgpt}.

\section{Conclusion}
\label{457883140213}
This survey has comprehensively examined the synergistic relationship between causality and deep generative models, each with its unique strengths and limitations in modeling data-generating processes. 
Furthermore, we delve into am emergent frontier of large-scale generative models, such as generative LLMs, elucidating their prospective role in advancing both methodology and application in causal research. 
Through a comprehensive review that spans methodologies, trustworthy properties and applications, this work not only fills a marked gap in existing literature but also serves as a foundational reference. 
By highlighting open challenges and suggesting prospective research directions, we hope to inspire continued exploration in this rapidly evolving area.



{
\footnotesize
\bibliographystyle{IEEEtran}
\bibliography{ref}
}

\vfill

\end{document}

%% file: secs/sec_Backgrond.tex
{
In this study, we adopt a holistic perspective that integrates causality with deep generative models (DGMs), treating them as complementary components of data generation processes. 
We explore the potential for synergy between the theoretical underpinnings of causality and the computational prowess of DGMs. 
Specifically, DGMs serve as sophisticated tools to model intricate, high-dimensional datasets, while causality provides the interpretable framework necessary for understanding the mechanisms that drive these data-generating processes. 
Thus, this section will delineate the foundational concepts, theoretical constructs, and methodological approaches that are pivotal to both domains.}

\subsection{Causality}
\label{004238198634}
{
Causality, as conceptualized by Judea Pearl, seeks to understand the cause-effect relationships that underlie data-generating processes \cite{pearl2009causal}. 
It goes beyond mere associations, which might arise from confounding factors, to decipher genuine causal mechanisms between variables. 
In this section, we explore the foundational concepts of SCMs and the ladder of causation, underscoring their significance in delineating the data-generating processes.
We also discuss the role of independent component analysis (ICA) in identifying latent causal factors.}

\subsubsection{Structural Causal Models}
Structural causal models (SCMs) provide a rigorous mathematical framework to represent causal relationships and explicitly characterize data-generating processes~\cite{pearl2009causality}. 
An SCM is denoted as \(\mathcal{M} := (\bm{S}, p(U))\), where \(\bm{S} = \{f^{(i)}\}_{i=1}^{K}\) signifies structural assignments:

\begin{equation}
\label{052652870031}
    \bm{x}^{(k)} := f^{(k)}(\bm{pa}^{(k)}; \bm{u}^{(k)})
\end{equation}
Here, \(\bm{x}^{(k)}\) is the \(k\)-th endogenous (or observed) variable. The term \(\bm{pa}^{(k)}\) denotes the parent set of \(\bm{x}^{(k)}\), representing its direct causal precursors. The joint distribution over independent exogenous noise variables is represented by \(p(U) = \prod_{i=1}^{K}p(\bm{u}^{(k)})\), with each noise variable, \(\bm{u}^{(k)}\), being uniquely associated with \(\bm{x}^{(k)}\). 

{For instance, consider a causal system with two variables, $\bm{x}^{(1)}$ and $\bm{x}^{(2)}$, represented by the following SCM:

\begin{align}
\bm{x}^{(1)} &= \bm{u}^{(1)}, \
&\bm{u}^{(1)} \sim \mathcal{N}(0, 1) \\
\bm{x}^{(2)} &= \frac{1}{1+\exp\{\bm{u}^{(2)}-\bm{x}^{(1)}\}} , \
& \bm{u}^{(2)} \sim \mathcal{N}(0, 1) 
\end{align}
Here, $\bm{u}^{(1)}$ and $\bm{u}^{(2)}$ are independent noise variables, following a normal distribution. This SCM exemplifies a basic causal relationship where $\bm{x}^{(1)}$ directly influences $\bm{x}^{(2)}$.

To illustrate a more intricate example, Figure \ref{040922319451} portrays a highly intricate and nonlinear interplay among variables within a dataset known as CelebA \cite{liu2015faceattributes}. In this dataset, variables encompass not only images but also various attributes such as age and gender. CausalGAN, a deep learning model, has been employed to infer the SCM from the CelebA dataset.
}

Central to SCMs is the notion that the joint distribution of endogenous variables can be articulated through \textit{causal mechanisms}, as described by \(p(\bm{x}^{(i)} \vert \bm{pa}^{(i)})\). 
This perspective contrasts with conventional factorizations and adheres to the \textit{Causal Markov condition}, as discussed in \cite{pearl2009causal,spirtes2000causation,scholkopf2021toward,Scholkopf2022FromST}:

\begin{equation}
\label{694530857668}
    p_{\mathcal{M}}(\bm{x}^{(i)},\cdots, \bm{x}^{(K)}) = \prod_{k=1}^{K} p(\bm{x}^{(k)} \vert  \bm{pa}^{(k)})
\end{equation}


\subsubsection{The Ladder of Causation}
\label{901357130845}

The \textit{Ladder of Causation} \cite{pearl2009causality,bareinboim2022pearl}, offers a structured way to understand the hierarchy of causal tasks.
It consists of three hierarchical layers, each representing a deeper level of causal reasoning.

The first layer, \textbf{\textit{"association"}}, focuses solely on observing statistical relationships, such as correlations. 
It doesn't involve any causal interpretations but merely observes patterns within data.
The second layer, \textbf{\textit{"intervention"}}, involves actively manipulating one or more variables to see the effects on others. 
This is often represented using the do-operator in causal models.
The third and most intricate layer is \textbf{\textit{"counterfactuals"}}. 
At this level, reasoning involves imagining alternative scenarios, often contrary to what actually happened. 
Questions like, "Would I have a headache had I not taken aspirin?" belong to this layer. 
Such counterfactual queries require a deep understanding of the underlying causal structure, making use of SCMs for precise formulations. 
The counterfactual layer is especially crucial for many advanced inference methodologies and will be elaborated upon in subsequent sections.

\subsubsection{Statistical tools for Causal Discovery}
\label{919611488502}
Causal discovery seeks to identify causal relations using empirical data \cite{Glymour2019ReviewOC}. Several statistical methodologies facilitate this task:
1) conditional independence relations in data help unveil causal structures, utilized by algorithms such as PC and FCI \cite{spirtes2000constructing}. 
2) functional causal models (FCMs) for continuous variables map direct causes to effects, with models like LiNGAM \cite{shimizu2006linear} and the PNL model \cite{zhang2006extensions, zhang2012identifiability} offering distinct approaches.
3) independent component analysis (ICA) transforms variables into independent components, initially crafted for signal separation. 
In causal discovery, methods like LiNGAM utilize ICA. 
With deep learning's rise, non-linear ICA has been harnessed for disentangled representation learning, with some works highlighting identifiability guarantees \cite{khemakhem2020variational, li2019identifying, kim2023covariate, xie2022multi, kong2022partial}.

\subsection{Deep Generative Models}
\label{306239698059}
Deep generative models (DGMs) harness the power of neural networks to capture intricate data distributions \cite{bond2021deep, wang2023scientific}. 
We delve into several conventional DGMs and touch upon the evolution towards large-scale generative frameworks, notably generative pretrained transformers \cite{radford2018improving}.

\subsubsection{Conventional DGMs}
Conventional DGMs in our survey primarily focus on foundational architectures that have paved the way for the development of current emerging large-scale generative models, which often encompass billions of parameters \cite{wang2023scientific}. 
These foundational architectures include GANs, VAEs, normalizing flows, and diffusion models \cite{bond2021deep}.

\textbf{Generative adversarial networks (GANs)}~\cite{Goodfellow2014GenerativeAN,Mirza2014ConditionalGA,Chen2016InfoGANIR,chen2021cde} consist of a generator and a discriminator competing in a two-player game. 
The generator aims to produce samples that are indistinguishable from real data, while the discriminator tries to distinguish between real and generated samples. 
The loss function is formulated as:
\begin{equation} 
\label{eq:gan}
\begin{aligned}
	\min \limits_{G} \max\limits_{D}\hspace{2pt}&\mathbb{E}_{\bm{x} \sim p(\bm{x})} \mathrm{log}[D(\bm{x})] + \mathbb{E}_{\bm{z} \sim p(\bm{z})} \mathrm{log}\left[1 - D(G(\bm{z}))\right].
\end{aligned}
\end{equation}
where \( \bm{z} \) is a noise vector typically sampled from a Gaussian distribution, and \( G(\bm{z}) \) is the generated sample.

\textbf{Variational autoencoders (VAEs)}~\cite{Goodfellow2014GenerativeAN,Mirza2014ConditionalGA,Chen2016InfoGANIR} are generative models that learn to encode and decode data in a probabilistic manner, consisting of an encoder and a decoder.
The encoder approximates the true posterior distribution of the latent variable given the data, and the decoder defines the likelihood of the data given the latent variable.
Specifically, the objective function of VAEs includes two terms: the reconstruction loss, which ensures that decoded data is similar to the original input and the KL-divergence term, which regularizes the encoder's output to follow a standard Gaussian distribution.
\begin{equation}
    \mathcal{L}_{\text{VAE}} = \mathbb{E}_{\bm{z} \sim q_{\phi}(\bm{z}|\bm{x})}[\log p_{\theta}(\bm{x}|\bm{z})] - D_{\text{KL}}(q_{\phi}(\bm{z}|\bm{x}) || p(\bm{z}))
\end{equation}
where \( q_{\phi} \) is the encoder's distribution and \( p_{\theta} \) is the decoder's distribution.

\textbf{Normalizing Flows}~\cite{dinh2016density,kingma2018glow,kobyzev2020normalizing} aim to transform a simple base distribution (e.g., Gaussian) into a complex distribution that resembles the data distribution.
They achieve this by applying a sequence o invertible transformations.
Each transformation in the sequence is chosen such that its Jacobian determinant is easy to compute, making it tractable to evaluate the data's density.

\textbf{Diffusion Models}~\cite{ho2020denoising,Song2021ScoreBasedGM} simulate a diffusion process to generate data. 
By establishing a Markov chain of diffusion steps, these models gradually introduce random noise to the data. 
The core idea is then to learn how to reverse this diffusion process, allowing the model to construct desired data samples from the introduced noise \cite{weng2021diffusion}.

\subsubsection{Large-scale DGMs}
In this review, large-scale DGMs primarily denote generative large language models (LLMs), with generative pretrained transformers (GPTs)~\cite{radford2018improving,radford2019language} being a prime example. 
These generative LLMs, characterized by their expansive architectures with often billions of parameters, have attracted significant attention in the machine learning domain. 
Noteworthy models include GPTs \cite{brown2020language,openai2023gpt4}, LLaMA \cite{touvron2023llama}, and Alpaca \cite{taori2023stanford}. 
Trained on extensive textual datasets, these models excel in diverse natural language understanding tasks, exhibiting intriguing properties such as scaling laws and emergent capabilities \cite{zhao2023survey}.
Furthermore, they have been investigated for the potential in causal tasks (\S \ref{605156580006}).

%% file: secs/sec_Causality4DGMs.tex
{
In this section, we explore the intersection of causal principles into DGMs.
Initially, we highlight the significance of integrating causality into DGMs (\S \ref{458711713679}) and proceed to define the scope and problem setting of causal deep generative models (CGMs) (\S \ref{430735514652}). Subsequently, we present a comprehensive taxonomy illustrating how causality is incorporated into DGMs (\S \ref{589253352307}), also outlined in Table \ref{302764778028}. 
This structured approach provides a clear understanding of the role and implications of causality in DGMs.
}

\begin{figure}[!t]
    \centering
    \includegraphics[width=0.95\linewidth]{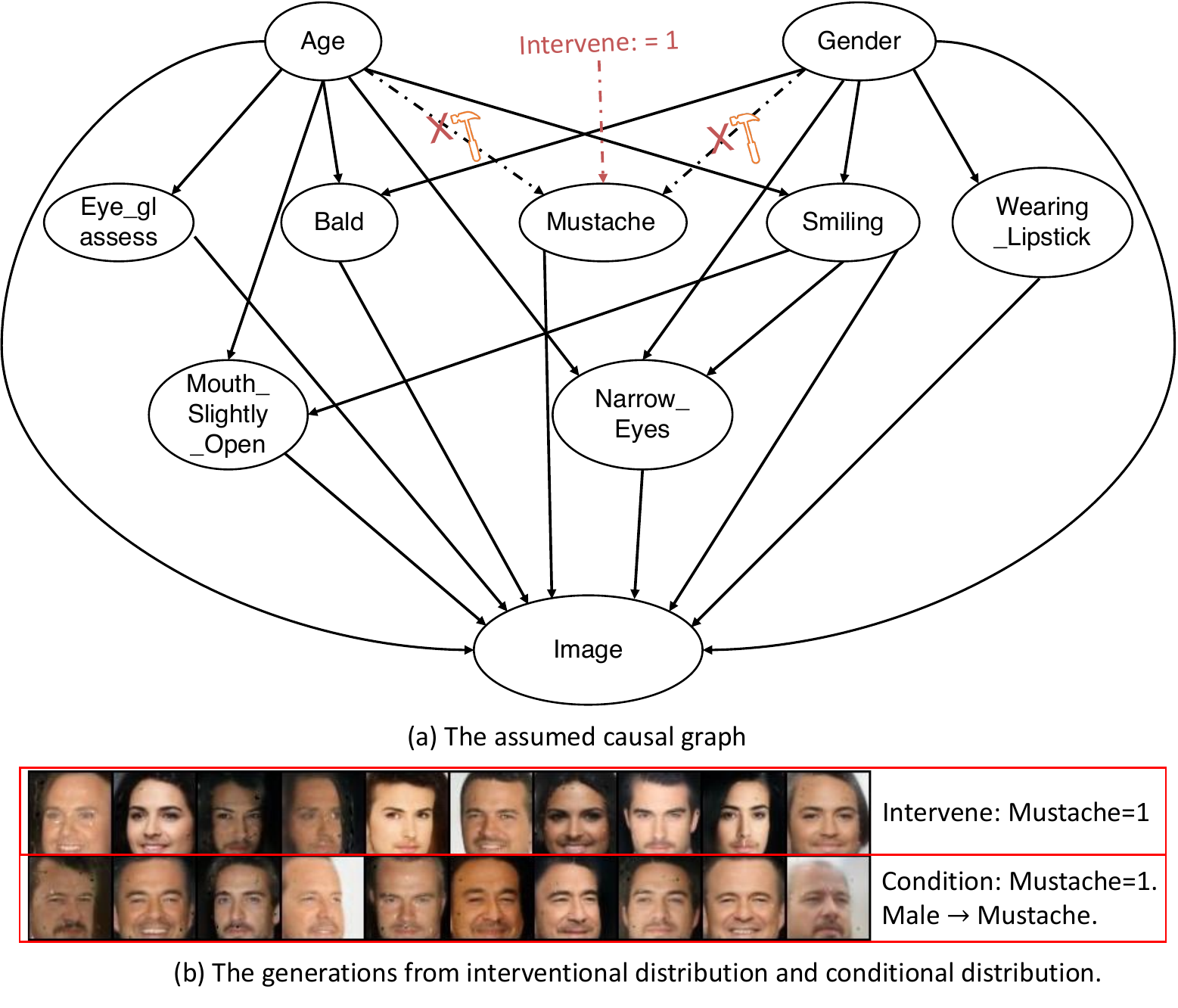} 
    \caption{In the CausalGAN framework \protect\cite{Kocaoglu2018CausalGANLC}, the depicted causal graph is based on a subset of the CelebFaces Attributes Dataset (CelebA) \protect\cite{liu2015faceattributes}. 
    This graph demonstrates how DGMs can isolate the causal influences of \(Age\) and \(Gender\) on the \(Mustache\) attribute when performing interventions. 
    Notably, both male and female faces appear when sampling from the interventional distribution with \(Mustache=1\), whereas only male faces are observed when sampled from the conditional distribution of \(Mustache=1\) since $P(Male=1\vert Mustache=1)=1$.}
    \label{040922319451}
\end{figure}

\subsection{Rationale for Causal Integration in DGMs}
\label{458711713679}
This subsection articulates two key advantages of incorporating causality into DGMs, aiming to resolve existing limitations and to significantly enhance their capabilities.

\subsubsection{Augmenting Extrapolative Capabilities for Controlled Generation}
DGMs are adept at simulating complex data distributions, facilitating the generation of samples with specific attributes. 
Nevertheless, they are confined by the observational distribution from which they sample, thereby limiting their extrapolative reach. 
By encapsulating the causal relationships among latent variables, causality extends the model's ability to generalize beyond the training distribution, thereby facilitating the generation of samples in new, unexplored contexts \cite{Besserve2021ATO,Kocaoglu2018CausalGANLC,xie2022multi}. 
For instance, as depicted in Figure \ref{040922319451}, intervening on the \(Mustache\) variable (while isolating it from causal factors like \(Age\) and \(Gender\)) enables the generation of facial images with the attributes \(\{Mustache=1, Gender=Female\}\) from the resulting interventional distribution. 
Without such intervention, conditioning on \(Mustache=1\) would absolutely yield images with \(Gender=Male\).
This capability has indispensable applications, notably in medical diagnostics and image editing, where generating out-of-distribution samples is often crucial \cite{de2018molgan,gregor2015draw,Mirza2014ConditionalGA,liu2019conditional}.

\subsubsection{Fortifying Interpretability via Causal Disentanglement} 
Interpretability in DGMs entails understanding the decisions made by models, making them transparent and trustworthy. 
While DGMs, notably VAEs, have shown promise in learning disentangled representations, which separate out distinct factors of variations in the data \cite{Bengio2013RepresentationLA, donahue2016adversarial, Higgins2017betaVAELB}, they typically struggle with nonlinearly mixed underlying factors \cite{Hyvrinen1999NonlinearIC, hyvarinen2019nonlinear}. 
Also, relying solely on statistical independence among latent factors often falls short in capturing the complex interactions and dependencies present in real-world data \cite{Locatello2019ChallengingCA, Besserve2020CounterfactualsUT}.
Causality introduces foundational SCMs and further advancements through the \textit{Independent Causal Mechanisms (ICM) Principle} \cite{Schlkopf2012OnCA, scholkopf2021toward} (\textit{modularity} and \textit{exogeneity} \cite{pearl2009causality}), presenting a more nuanced approach.
It not only focuses on the separation of factors but also ensures that these factors have meaningful causal relationships, enhancing the overall interpretability \cite{scholkopf2021toward, kong2022partial}.
For example, consider an illustrative facial image in Figure \ref{040922319451}: while a conventional DGM might identify features such as "eyes" and "mouth" independently, a causally-disentangled model would recognize that the attribute of \(Smiling\underline{\hspace{0.5em}}\) is more likely to result in \(Mouth\underline{\hspace{0.5em}}Slightly\underline{\hspace{0.5em}}Open\) and \(Narrow\underline{\hspace{0.5em}}Eyes\) \cite{Kocaoglu2018CausalGANLC,shen2022weakly}.
Such nuanced insights are particularly invaluable in fields like medical diagnostics, where understanding the intricate relationships between features can lead to more accurate and actionable results.

\subsection{Setting the Stage for Causal Generative Models}
\label{430735514652}
In this work, we offer an explicit definition and scope for causal deep generative models (CGMs), as shown in Definition \ref{566776443147}.
In contrast to conventional DGMs, which primarily focus on approximating complex data distributions, CGMs are engineered to model the underlying causal mechanisms that govern these distributions. 
This leads to enhanced capabilities in terms of robustness, interpretability, and the ability to undertake meaningful interventions and counterfactual analyses \cite{zhang2020causal}.

\begin{definition}(Causal Deep Generative Models)
\label{566776443147}
Causal deep generative models (CGMs) are a specialized subclass of deep generative models (DGMs).
They stand out by incorporating causal structures in their model design, which influences both the architecture and objective functions.
While some CGMs operate by introducing latent variable $z$ that is designed to be manipulatable, reflecting potential causal relationships, others may adapt SCMs directly into their architecture or leverage DGMs to implement interventions.
Formally, while a typical DGM is represented as $p(\bm{x}\vert \bm{z}; \theta)$, a CGM extends this by integrating an explicit causal structure $C$--either over $\bm{z}$ or $\bm{x}$, within its architecture, or as an intervention mechanism based on SCMs, resulting in causally-informed distribution $p(\bm{x}\vert \bm{z}, C; \theta^{'})$. 
\end{definition}

\subsection{Classifying Methods of Causal Integration}
\label{589253352307}
This subsection presents a taxonomy of representative CGMs, categorizing them based on the types of generative architectures they employ, namely GANs-based, VAEs-based, and Diffusion-based models. Special emphasis is placed on elucidating the various methodologies for effectively incorporating causal principles into these DGMs.

\subsubsection{GANs-Based Models}
Despite GANs' impressive performance in data generation tasks \cite{Mirza2014ConditionalGA, Goodfellow2014GenerativeAN}, controllability and interpretability remain challenges, as discussed in $\S$ \ref{458711713679}. 
To address these issues, recent works integrate causality into GANs, either by mapping explicit SCMs within architectures or by incorporating causal mechanisms into GAN's generative strategies.

\begin{figure}[!t]
    \centering
    \includegraphics[width=\linewidth]{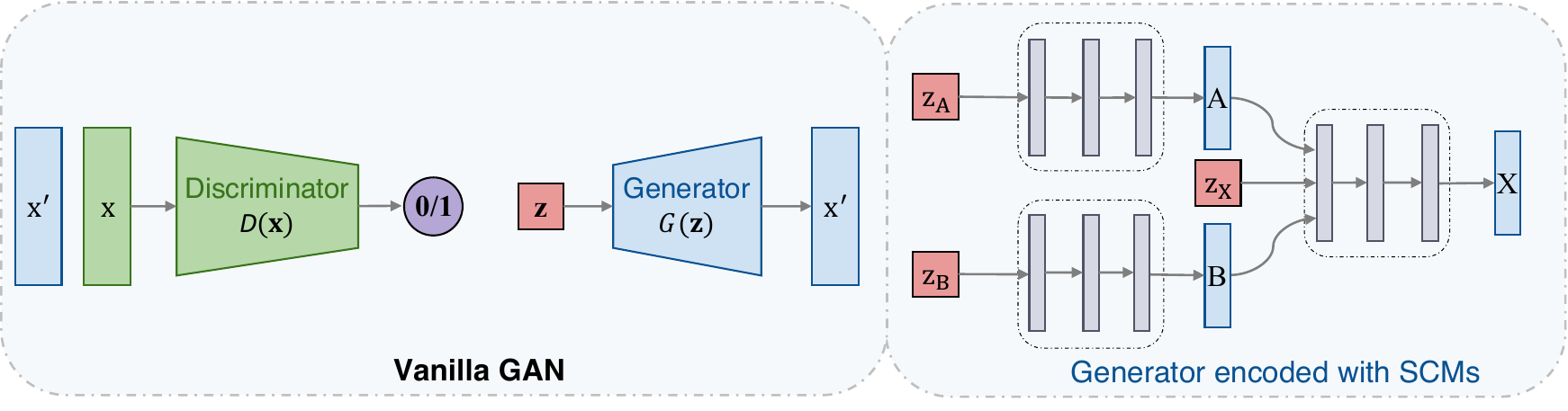} 
    \caption{CausalGAN \protect\cite{Kocaoglu2018CausalGANLC} illustrates the process of translating an SCM into the structure of a generator architecture  using a simplified causal graph $\{A\rightarrow X \leftarrow B\}$ \cite{weng2021diffusion}.}
    \label{925818866720}
\end{figure}

\paragraph{\textbf{SCMs as a Generator in GANs}}
\cite{Kocaoglu2018CausalGANLC} pioneered the integration of SCMs into GANs with their work on CausalGAN. 
The generator architecture in CausalGAN is uniquely designed based on an assumed causal graph, as illustrated in Figure \ref{925818866720}. 
To provide a concrete example, consider two attributes, ${Gender}$ and ${Mustache}$, from the CelebA dataset \cite{liu2015faceattributes}. 
A causal relationship can be described as $\{Age \rightarrow Mustache \leftarrow Gender\}$, depicted in Figure \ref{040922319451}. 
When sampling from the conditional distribution $P(\cdot \vert Mustache=1)$, the data primarily comprises males, as ${Gender=Male}$ causally influences ${Mustache=1}$.

However, the beauty of integrating causality lies in its ability to support intervention on causal variables within the SCM framework. 
As shown in Figure \ref{040922319451}, intervening on the $Mustache$ variable involves modifying its structural assignments, typically through atomic interventions, setting $Mustache$ to one or zero. 
This disentangles $Mustache$ from its causal precursors, $\{Age, Gender\}$, thereby nullifying their causal effects. 
Consequently, the model can generate samples that are not present in the training distribution, such as females or children with mustaches, extending the model's generalization capabilities beyond existing conditional generation methods.

While CausalGAN offers theoretical guarantees for accurate sampling from interventional distributions, its applicability is currently limited to binary attributes. Furthermore, despite its innovative contributions to generating out-of-distribution samples, the utility of such generated data for downstream tasks remains an open question \cite{Kocaoglu2018CausalGANLC}.

In a vein similar to CausalGAN, the Counterfactual Generative Network (CGN) \cite{Sauer2021CounterfactualGN} incorporates SCMs into the generator's architecture to control specific factors of variation, such as shape, texture, and background. 
The generator function in CGN is formally described as:
\begin{equation}
    \label{873045624885}
    \bm{x}_{\text{gen}} = C(\bm{m}, \bm{f}, \bm{b}) = \bm{m} \odot \bm{f} + (1-\bm{m}) \odot \bm{b}
\end{equation}
where \(\bm{m}\) represents the mask, \(\bm{f}\) is the foreground, and \(\bm{b}\) stands for the background. The operator \(\odot\) signifies elementwise multiplication. 
CGN, unlike CausalGAN, aims for robust, interpretable classifiers and mitigates shortcut learning \cite{Geirhos2020ShortcutLI}. 
Using counterfactuals for data augmentation, it seeks classifiers invariant to data variations. 

\cite{Wen2021CausalTGANGT} introduced Causal-TGAN for tabular data generation, embedding predefined causal relationships into GANs for improved data accuracy. 
On the fairness front, \cite{Xu2019AchievingCF} proposed CFGAN to enforce causal fairness principles, including total effect and direct/indirect discrimination. 
Additionally, \cite{Breugel2021DECAFGF} designed DECAF, a GAN that integrates SCMs into the generator's input layers for generating fair synthetic data, capturing causal mechanisms essential for fairness.


\begin{figure}[!t]
    \centering
    \includegraphics[width=\linewidth]{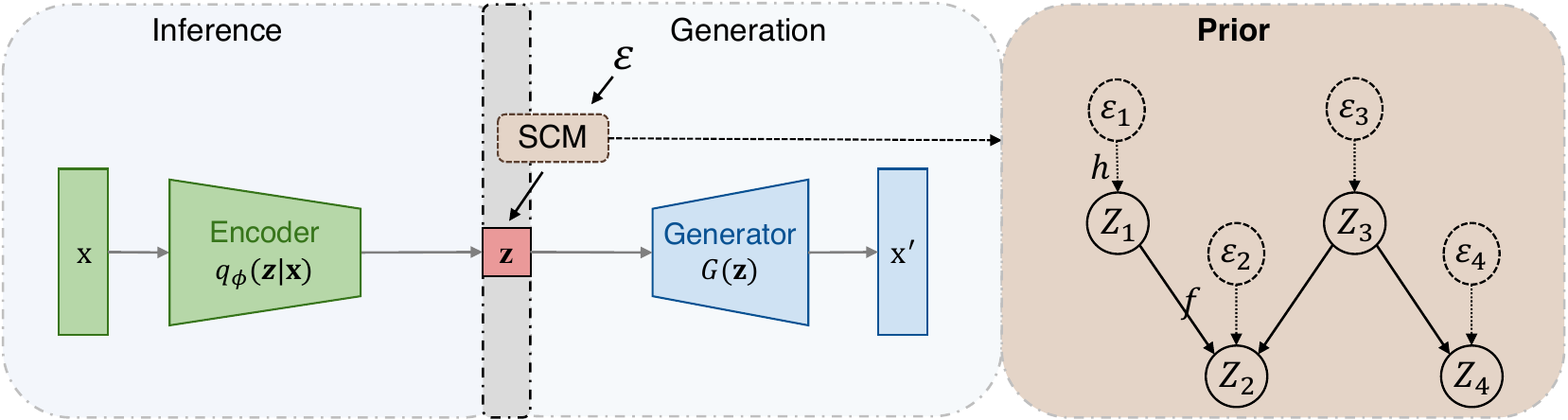} 
    \caption{The illustration of encoding an SCM as the prior for latent variables in bidirectional generative models \protect\cite{shen2022weakly, weng2021diffusion}.}
    \label{071106881268}
\end{figure}

\paragraph{\textbf{SCMs as a Prior for Latent Variables}}
Shen \etal~\cite{shen2022weakly} integrate an explicit causal structure $C$ over latent variable $\bm{z}$ in a bidirectional generative model, and present an alternative: Disentangled gEnerative cAusal Representation (DEAR), as illustrated in Figure \ref{071106881268}. 
The model employs a weighted adjacency matrix $\bm{A}$ to represent the directed acyclic graph (DAG) upon the $k$ elements of $\bm{z}$ (i.e., four elements in Figure \ref{071106881268}), and simultaneously learns both the causal structure and the structural assignments via general nonlinear functions $f$ and $h$.
\begin{equation}
    \label{216591362718}
    \bm{z} = f(({\bm{I}-\bm{A}^{\top}})^{-1} h(\bm{\epsilon}))
\end{equation}
Here, $\bm{A}$ is the weighted adjacency matrix, and $\epsilon$ represents exogenous noise variables following $\mathcal{N}(\bm{0},\bm{I})$. 
$f$ and $h$ are element-wise nonlinear transformations.
In this manner, the causal relationships between variables are captured through the weighted adjacency matrix $\bm{A}$. 

Notably, the supervision method employed in \cite{shen2022weakly} is both more direct and arguably stronger than that in \cite{Khemakhem2020VariationalAA}, which utilizes time or domain indices for supervision. 
To clarify, \cite{shen2022weakly} necessitates strong supervision of annotated labels corresponding to the true underlying factors, ensuring the identifiability of the latent causal variables. 
{iStyleGAN \cite{xie2022multi} proposes a generative process for multi-domain image and show the identifiability of latent variables with observed domain indices, effectively generating paired samples from unpaired data. }



\paragraph{\textbf{Generative Intervention}}
Rather than incorporating SCMs directly into GANs, alternative works like \cite{Mao2021GenerativeIF} and \cite{Zhang2021LearningCR} leverage GANs to implement SCM-based interventions for downstream tasks. 
GenInt \cite{Mao2021GenerativeIF} aims to enhance the robustness of visual classifiers by using a conditional GAN (cGAN) \cite{Mirza2014ConditionalGA} to intervene on spurious variables, effectively performing adversarial augmentation constrained by causal relations. 
This results in improved model performance by eliminating spurious correlations. Meanwhile, \cite{Zhang2021LearningCR} applies a similar interventionist approach to 3D pose estimation, considering domain shifts as interventions. They use deep generative models to simulate these shifts, leading to more transferable and causally-informed representations that boost estimation accuracy.

\paragraph{\textbf{Knowledge Extrapolation}}
Addressing the challenge of incorporating structural causal elements directly into the generator architectures of GANs, Feng \etal~\cite{Feng2022PrincipledKE} introduce PKD.
This approach focuses on generating high-fidelity counterfactual samples using pre-trained, state-or-the-art generators. 
The method posits that the extrapolation distribution primarily deviates from the original in the dimension specific to the extrapolated knowledge.
This divergence is adeptly modeled through an adversarial framework, ensuring for seamless integration with cutting-edge GANs.

\subsubsection{VAEs-Based Models}
VAEs are renowned for their disentangled latent representations and stand as a state-of-the-art in deep generative models \cite{Kingma2014AutoEncodingVB, Higgins2017betaVAELB}. While many studies assume the latent variables in VAEs to be mutually independent, this notion is challenged by insights from nonlinear Independent Component Analysis (ICA). Specifically, the generative function is generally unidentifiable in nonlinear settings. However, recent advancements \cite{Hyvrinen1999NonlinearIC, hyvarinen2019nonlinear, Khemakhem2020VariationalAA, kong2022partial} have shown that the function becomes identifiable when multiple data distributions share the same generative function but differ in the distributions of their latent variables.

\begin{figure}[!tb]
    \centering
    \includegraphics[width=0.48\textwidth]{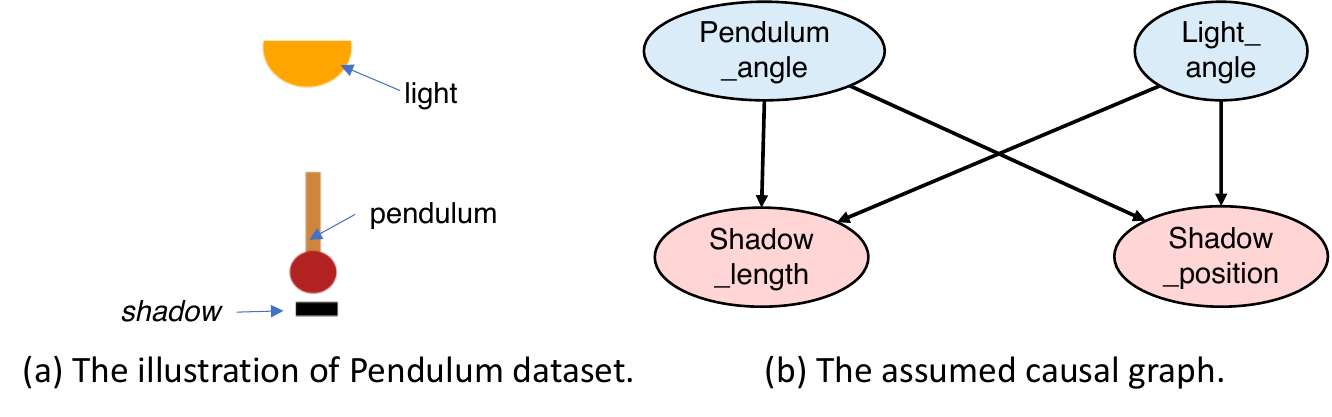} 
        \caption{The illustration and presumed causal graph for the Pendulum dataset~\cite{Yang2021CausalVAEDR}. 
        }
        \label{019405822627}
\end{figure}

\paragraph{\textbf{SCMs Integrated in VAEs}}

\begin{figure}[!tb]
    \centering
    \includegraphics[width=\linewidth]{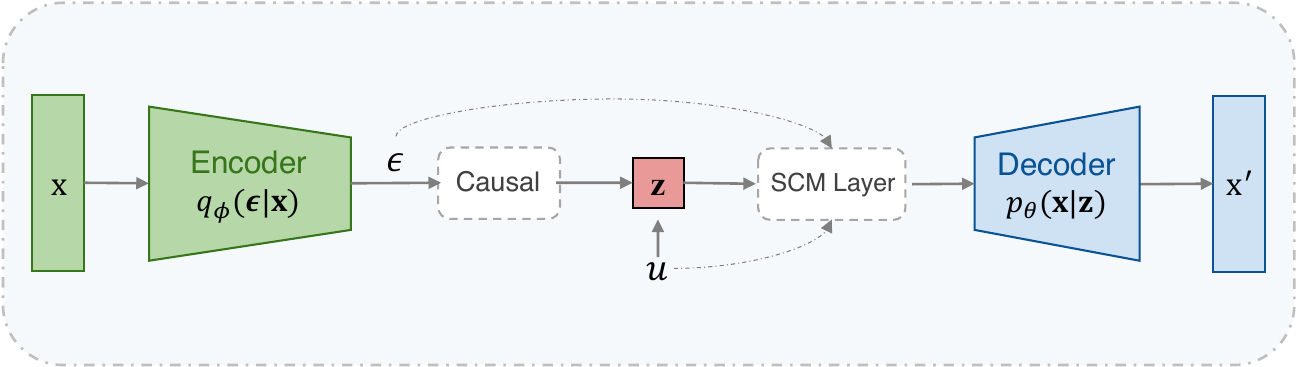} 
    \caption{
    As outlined in \protect\cite{Yang2021CausalVAEDR}, the architecture of CausalVAE extends a vanilla VAE by incorporating a causal module and an SCM layer. 
    The causal module employs an adjacency matrix $\bm{A}$ to convert independent exogenous variables, denoted by $ \bm{\epsilon} $, into their corresponding causal representations. The SCM layer subsequently manipulates these causal representations $ \bm{z} $, to simulate the propagation of causal effects. Furthermore, supervision is provided for the variables of interest through $ \bm{u} $.}
    \label{122145664592}
\end{figure}

While \cite{Locatello2019ChallengingCA} define disentangled representation as isolating unique, semantic factors of variation in data, Yang \etal~\cite{Yang2021CausalVAEDR} challenge this by asserting that these factors can also be causally interdependent, exemplified by how changes in a light source or pendulum position affect shadow characteristics, as shown in Figure \ref{019405822627}.  
To address this complexity, \cite{Yang2021CausalVAEDR} introduces CausalVAE, which embeds an SCM into a standard VAE architecture, as shown in Figure \ref{122145664592}.
CausalVAE integrates a causal component and an SCM layer into the standard VAE architecture. 
This component utilizes an adjacency matrix $\bm{A}$ to represent the causal DAG, as described by Eq. (\ref{932426115568}): 
\begin{equation}
    \label{932426115568}
    \bm{z} = \bm{A}^{T}\bm{z} + \bm{\epsilon} = (I-\bm{A}^T)^{-1} \bm{\epsilon}, \bm{\epsilon} \sim \mathcal{N}(\bm{0}, \bm{I})
\end{equation}
CausalVAE uses key data concepts as supervision signals, such as attributes in the Pendulum dataset (Figure \ref{019405822627}). 
It employs a conditional prior $p(\bm{z} \vert \bm{u})$ on these concepts to regularize the learned posterior of $\bm{z}$, unlike iVAE \cite{khemakhem2020variational} which uses weaker forms of supervision.
The SCM layer facilitates interventions through a Mask Layer \cite{ng2022masked}, modifying the learned adjacency matrix $\bm{A}$ to convey causal relationships in latent variables.

Importantly, the identifiability in CausalVAE is guaranteed through stronger supervision on the latent causal variables \cite{Khemakhem2020VariationalAA} with explicit annotations, as opposed to the weaker time or domain-based indices in \cite{khemakhem2020variational}. 

{
iVAE \cite{khemakhem2020variational} adopts a factorized prior distribution based on an observed variable, ensuring identifiable latents.
CI-iVAE \cite{kim2023covariate} highlights iVAE's potential "posterior collapse" issue, where the impact of $\bm x$ diminishes, leading to $q(\bm z|\bm x,\bm u)=p(\bm z|\bm u)$. To tackle this, CI-iVAE suggests deriving the latent solely from $\bm x$ and combining $q(\bm z|\bm x)$ and $q(\bm z|\bm x, \bm u)$ to prevent posterior collapse. iMSDA~\cite{kong2022partial} introduces domain index as an additional variable, extracting invariant components for domain adaptation. The methods \cite{khemakhem2020variational,kim2023covariate,kong2022partial} rely on the auxliary variable to identify the true latents. Interestingly, \cite{willetts2021don} show that strong inductive bias on the latents (by empliying a Gaussian mixture VAE \cite{jiang2016variational}) can achieve similar identifiability results compared with the supervised one \cite{khemakhem2020variational} empirically. \cite{kivva2022identifiability} formally shows that the latents can be recovered with data only with the mixture prior in VAE.}


\paragraph{\textbf{Estimate SCMs with VAEs for Generation}}
Rather than directly incorporating SCMs, various works have leveraged VAEs to infer structural assignments \cite{Pawlowski2020DeepSC, Reinhold2021ASC, SnchezMartn2021VACADO}. 
Among these, \cite{Pawlowski2020DeepSC} introduces DeepSCM, a deep generative framework designed to deduce counterfactual effects using a well-defined causal graph. 
Utilizing normalizing flow techniques in conjunction with a VAE, the authors estimate both the exogenous noise variables linked to endogenous variables and their structural relationships. 
This comprehensive SCM facilitates a triad of causal procedures: abduction, action, and prediction, offering an innovative means of addressing causal inquiries. 
The framework is further applied to brain MRI imaging, yielding intriguing, albeit qualitatively assessed, counterfactual alterations. 
It's worth noting that the methodology cannot account for latent confounders and assumes causal sufficiency \cite{Scholkopf2022FromST}. 
Besides, \cite{ribeiro2023high} presents pragmatic causal generative modelling framework for estimating high-fidelity image counterfactuals using deep conditional Hierarchical Variational Autoencoders (HVAEs).

\cite{SnchezMartn2021VACADO} uses graph neural networks (GNNs) and variational graph auto-encoders (VGAE) to encode SCMs and approximate interventional effects. 
The model assumes a known causal graph and no hidden confounders, but its GNNs-based message aggregation may limit its ability to handle complex causal structures.
\cite{Hu2021ACL} shifts the emphasis in text generation towards a causal framework. 
Focusing on controllable text generation, they aim to produce text with specific attributes and modify existing samples accordingly. 
Analogous to DeepSCM \cite{Pawlowski2020DeepSC}, they employ VAEs to deduce exogenous variables and structural relationships among causal variables, albeit across diverse application domains.

\begin{table*}[htb]
    \centering 
    \caption{Summary of causal discovery approaches through DGMs.}
     \begin{tabular}{>{\centering\arraybackslash}p{1.8cm}>{\centering\arraybackslash}p{2.6cm}>{\centering\arraybackslash}p{2.6cm}>{\centering\arraybackslash}p{4.8cm}>{\centering\arraybackslash}p{1.8cm}>
     {\centering\arraybackslash}p{1.8cm}}
        \toprule
        \textbf{Model} & \textbf{DGMs}& \textbf{Goal} & \textbf{Key Assumption} & \textbf{MultiVariate}  & \textbf{Category} \\ \midrule
         CAREFL \cite{khemakhem2020ice} & Normalizing Flows & Causal Ordering & Affine Generation Process& \xmark & Combinatoric \\ 
         CAN \cite{moraffah2020causal} & GANs & Causal Graph & LinearSCM & \checkmark & Continuous \\ 
        OCDaf \cite{kamkari2023ocdaf} & Normalizing Flows & Causal Ordering & Invertible Location-Scale Noise Model & \checkmark & Continuous \\ 
        DiffAN \cite{sanchez2022diffusion} & Diffusion Models & Causal Ordering & Additive Noise Model  & \checkmark  & Continuous\\ 
        SAM \cite{kalainathan2022structural} & GANs & Causal Graph &  Markov Kernel & \checkmark & Continuous \\ \bottomrule
    \end{tabular}
    
    \label{tab:dgm_discovery}
\end{table*}

\subsubsection{Diffusion-Based Models}
\label{831281179594}

Recent studies have investigated the integration of structural causal models with generative diffusion models for counterfactual estimation and explanation \cite{Dhariwal2021DiffusionMB}. 
This interdisciplinary approach has garnered attention in various works \cite{Sanchez2022DiffusionCM,Augustin2022DiffusionVC}.

\begin{figure}[!b]
    \centering
    \includegraphics[width=\linewidth]{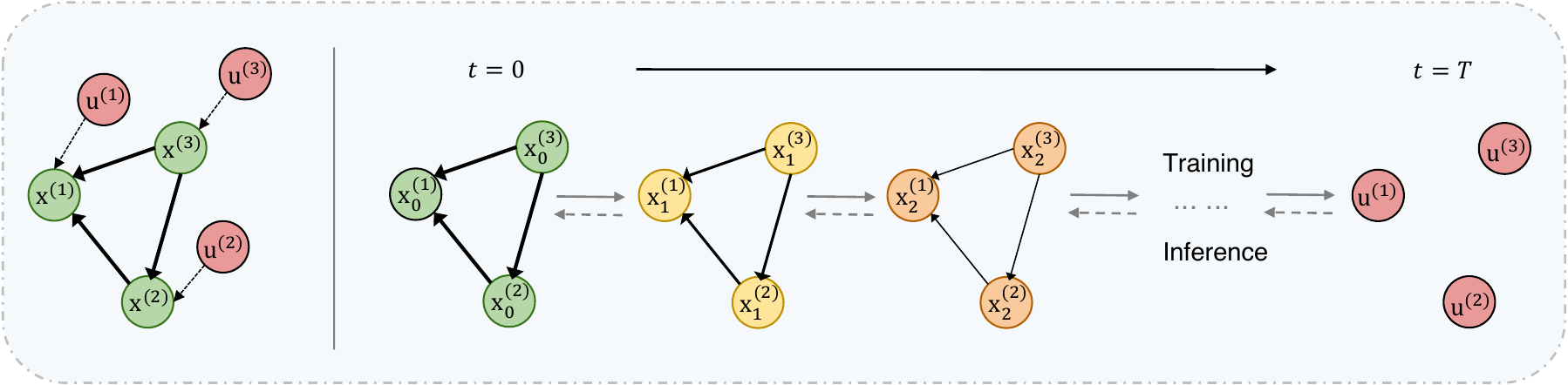} 
    \caption{ The Diff-SCM model, which is a structured generative model, illustrates how the diffusion process can weaken the causal relationships between endogenous variables that are causally linked \protect\cite{Sanchez2022DiffusionCM}.}
    \label{802199783739}
\end{figure}

\cite{Sanchez2022DiffusionCM} proposes Diff-SCM as a way to unite SCMs and the stochastic differential equation (SDE) framework \cite{Song2021ScoreBasedGM,weng2021diffusion}.
As depicted in Figure \ref{802199783739}, they formulate a forward diffusion as a process of weakening the causal relations between endogenous variables.
Specifically, the original joint distribution entailed by the Eq. (\ref{694530857668}) that diffuses to independent Gaussian distributions $p(U)$, as  $p(\mathbf{x}_{t=T}^{(k)}) = p(\mathbf{u}^{(k)})$ and $p(\mathbf{x}_{t=0}^{(k)}) = p_{\mathcal{M}}(\mathbf{x}^{(k)})$ with $t \in [0, T]$.
Following the time-dependent SDE \cite{Song2021ScoreBasedGM}, Diff-SCM is defined as:
\begin{equation}
\begin{aligned}
    \label{143892833217}
    d\mathbf{x}^{(k)} &= -\frac{1}{2} \beta_{t} \mathbf{x}^{(k)} dt + \sqrt{\beta_t}d\mathbf{w}, \forall k \in [1, K], \\
    \text{where}\   p(&\mathbf{x}_{0}^{(k)}) = \prod_{j=k}^{K} p(\mathbf{x}^{(j)} \vert \mathbf{pa}^{(j)}), p(\mathbf{x}_T^{(k)}) = p(\mathbf{u}^{(k)})
\end{aligned}
\end{equation}
where $\beta_t \in [0,1]$ represents the time-dependent variance of a Gaussian noise introduced in each forward process and $\mathbf{w}$ is the Brownian motion leading to a Gaussian distribution.  

This study utilizes an anti-causal predictor inspired by the classifier-guided diffusion introduced in previous work \cite{Dhariwal2021DiffusionMB}.
This approach involves transforming the prior distribution $p(U)$ into the data distribution through a gradual removal of noise using the gradient of the data distribution, until $\mathbf{x}_{t=0}^{(k)} = \mathbf{x}^{(k)}_{}$ (as described in \cite{Song2021ScoreBasedGM}).
With the aim of achieving conditional generation, \cite{Dhariwal2021DiffusionMB} train a classifier and use its gradients to guide the diffusion process towards the desired conditioning information.
To obtain counterfactual samples, they pass the gradients of the predictor through a reverse-diffusion process after performing an intervention (assigning a specific value to the cause).
These counterfactual samples are highly similar images that differ only in their class label.
This work follows three steps of causal hierarchy, similar to \cite{Pawlowski2020DeepSC}, but is limited to a bi-variable SCM with the causal relationship $\mathbf{x}^{(2)} \rightarrow \mathbf{x}^{(1)}$, where $\mathbf{x}^{(1)}$ represents an image, $\mathbf{x}^{(2)}$ represents a class label.

\cite{chao2023interventional} presents Diffusion-based causal models (DCM), which employs diffusion models to craft unique latent encodings that facilitate causal inference through direct sampling under interventions and counterfactual reasoning. 
Meanwhile, \cite{Sanchez2022WhatIH} extends Diff-SCM to create the counterfactual diffusion probabilistic model (CDPM), which incorporates implicit attention for lesion segmentation. 
This approach generates healthy counterfactual images from patient inputs, utilizing the differences to construct pathology heatmaps and achieving superior performance over traditional downstream classifiers. 
Further research continues to explore the application of diffusion models in generating counterfactual images and providing explanations \cite{Goyal2019CounterfactualVE,Augustin2022DiffusionVC,bedel2023dreamr}.

%% file: secs/sec_DGMs4Causality.tex



\begin{table*}[thb]
 \caption{Summary of counterfactual inference leveraging DGMs.}
 \begin{tabular}{>{\centering\arraybackslash}p{2cm}>{\centering\arraybackslash}p{3cm}>{\centering\arraybackslash}p{2.6cm}>{\centering\arraybackslash}p{2.3cm}>{\centering\arraybackslash}p{2.3cm}>
 {\centering\arraybackslash}p{3.3cm}}
    \toprule
        \textbf{Model} & \textbf{DGMs}& \textbf{Noise Estimation} & \textbf{MultiVariate} & \textbf{Continuous Intervention} & \textbf{Additional Information} \\ \midrule
        CAREFL \cite{khemakhem2021causal} & Normalizing Flows & Flow Inversion & \checkmark & \checkmark & Causal Ordering \\ 
        DeepSCM \cite{Pawlowski2020DeepSC} & Normalizing Flows  & Amortised Variational Inference & \checkmark & \checkmark & Causal Graph \\ 
        \multirow{2}{*}{BGM \cite{nasr2023counterfactual}} & \multirow{2}{*}{Normalizing Flows} & \multirow{2}{*}{Flow Inversion} & \multirow{2}{*}{\checkmark} &\multirow{2}{*}{\checkmark} & Instrumental\\
        &&&&&or Backdoor Variable \\ 
        Diff-SCM \cite{sanchez2022diffusion} & Diffusion Models & DDIM Inversion & \xmark &\checkmark  & --  \\ 
        DCM \cite{chao2023interventional} & Diffusion Models & DDIM Inversion & \checkmark &\checkmark& -- \\ 
        GANITE \cite{yoon2018ganite} & GANs & \xmark & \checkmark &\xmark& -- \\   
         SCIGAN \cite{bica2020estimating} & GANs & \xmark & \checkmark &\checkmark& -- \\ \bottomrule
    \end{tabular}
   
    \label{tab:dgm_cfr}
\end{table*}

{
In this section, we shift our focus from the integration of causality in DGMs, as discussed in \S \ref{971699495725}, to the specific application of DGMs in identifying causal relationships.
While \S \ref{971699495725} covers how causality principles enhance DGMs' capabilities in generation and classification tasks, here we explore how DGMs can be effectively used for causal discovery, inference and representation learning.
This section delves into the emerging potential of DGMs to unravel complex causal structures.
}
\subsection{Causal Discovery through DGMs}
\label{285171066616}

Causal information is essential for numerous scientific and engineering tasks. However, conducting randomized experiments to determine causal relationships among observed variables can be costly and challenging. As a result, causal discovery, which infers these relationships, has become invaluable. Lately, the use of deep generative models (DGMs) for causal discovery has gained popularity. 
A summary of these approaches can be found in Table \ref{tab:dgm_discovery}.

Causal discovery methods can be broadly divided into combinatoric and continuous optimization approaches \cite{vowels2022d}. Combinatoric methods explore the structure space, selecting models based on specific criteria like conditional independence or optimal score functions. 
The CAREFL approach \cite{khemakhem2021causal}, for instance, frames causal discovery as a statistical testing challenge, drawing inspiration from \cite{hyvarinen2013pairwise}. 
For bi-variate scenarios, CAREFL models the generation of one variable using an autoregressive flow, influenced by a parent variable. 
The method then determines the causal ordering from candidates based on log-likelihood scores from a validation dataset. 
Alternatively, GCIT \cite{bellot2019conditional} employs GANs to model the generation of $X$ from $Z$. By comparing generated and real samples, it performs a statistical test to verify the conditional independence $X\perp Y | Z$, aiding the causal discovery process.

In contrast, continuous optimization-based approaches learn the structure directly from data. The CAN method \cite{moraffah2020causal} is designed to ascertain this structure and generate samples based on it using GANs. 
It operates under the assumption of a linear structural causal model in the latent space and enforces the DAG-constraint \cite{zheng2018dags} on the linear mixing matrix. 
Consequently, CAN can produce interventional samples.
DiffAN \cite{sanchez2022diffusion} employs diffusion to learn the score function, computing the Hessian entries via back-propagation. 
It then establishes the causal ordering by selecting the variable with the lowest variance as the leaf node, a strategy inspired by \cite{rolland2022score}. 
OCDaf \cite{kamkari2023ocdaf} extends CAREFL to handle multi-variate scenarios and introduces a continuous search algorithm for causal discovery using autoregressive flows. 
It incorporates a differentiable proxy loss, which combines various loss functions under different causal orderings. 
The dominant causal ordering emerging post-training is then selected.
In situations involving three variables, such as $X, Y,$ and $Z$, SAM \cite{kalainathan2022structural} suggests generating one variable from the other two using GANs. 
A learnable mask within this process aids in the selection of parent variables.

\subsection{Counterfactual Inference Leveraging DGMs}
\label{722408981966}
Counterfactual inference, positioned at the third level of Pearl's causal hierarchy, addresses questions like, "What would have happened to the patient if she had received a different treatment?". 
However, this domain presents challenges due to the absence of data for counterfactual scenarios. 
While there exist effective non-DGMs methods for counterfactual inference \cite{johansson2016learning, de2022deep}, this section emphasizes the role of DGMs in enhancing counterfactual inference. 
A summary of our discussion can be found in Table \ref{tab:dgm_cfr}.

Given factual observations $\langle X,Y \rangle$, where $Y = f(X, \epsilon)$ with $\epsilon$ as noise, $X$ as the covariate, and $Y$ as the outcome, the goal of counterfactual inference is to determine the value of $Y$ if $X$ were $X'$. This process can be broken down into three steps as outlined by \cite{pearl2018book}: 
1) Abduction: Use the factual observations to estimate the noise term $\epsilon$.
2) Action: Update the causal model by setting $X$ to its new value $X'$.
3) Prediction: Infer the counterfactual outcome using the updated model.

In most counterfactual inference scenarios, DGMs aim to enhance the estimation of the noise term during the abduction step \cite{pawlowski2023answering,nasr2023counterfactual,sanchez2022diffusion, chao2023interventional,khemakhem2020ice}. 
Some, however, focus on aligning the distributions of observations to perform regression \cite{yoon2018ganite,bica2020estimating}. 
The DeepSCM \cite{Pawlowski2020DeepSC} method provides an example of the former approach. 
It models the structural causal model using normalizing flows and captures relationships between variables. 
DeepSCM deploys a non-invertible deep neural network to derive semantic representations from noise and parent variables. 
These representations then parameterize the invertible conditional normalizing flows. 
After optimizing the likelihood's lower bound, DeepSCM determines the estimated noise for each variable by reversing the normalizing flow. It then approximates the counterfactual distribution using Mont Carlo methods. 
BGM~\cite{nasr2023counterfactual} establishes that counterfactual outcomes are identifiable when the SCM $f$ is monotonic concerning the noise term $\epsilon$. 
The paper introduces the use of conditional spline flow to simulate the generation process. By reversing this flow, counterfactual inference is performed. 
Additionally, BGM identifies counterfactual outcomes in the presence of instrumental variables or those adhering to the backdoor criterion. 
DCM~\cite{chao2023interventional} handles causal graphs with multiple variables. CAREFL~\cite{khemakhem2021causal} posits a causal ordering for variables and models them with auto-regressive flows. 
By reversing this flow, the noise is derived.

Unlike those methods which estimate the noise terms and perform counterfactual inference, 
GANITE \cite{yoon2018ganite} first learns a counterfactual generator in GAN by matching the joint distribution of observed covariate and outcome variables. 
Then it generates a dataset by feeding different treatment values and random noises and learns a individual treatment effects (ITE) generator to predict the factual and counterfactual outcomes directly. 
Based on GANITE, SCIGAN \cite{bica2020estimating} proposes a hierarchical discriminator to learn the counterfactual generator when interventions are continuous, e.g., the dosage of the treatment.

\subsection{Causal Representation Learning using DGMs}
\label{147110379187}
{Traditional causal discovery and inference assume that the observed random variables are connected by a causal graph \cite{scholkopf2021toward}. However, the observations may not be well-structured and causal representation learning seeks to extract genuine low-dimensional latent variables from the complex high-dimensional data. DGMs have been demonstrated to be beneficial in causal representation learning, such as VAE \cite{khemakhem2020variational,kim2023covariate,kong2022partial,Yang2021CausalVAEDR} and GAN \cite{shen2022weakly,xie2022multi} as we mentioned in \S \ref{971699495725}. 
Recently, there have been some representation learning methods with diffusion model \cite{preechakul2022diffusion,wang2023infodiffusion,zhang2022unsupervised,abstreiter2021diffusion,mittal2023diffusion,hudson2023soda}. 
\cite{wang2023infodiffusion} proposes to maximize the mutual information between the encoded latents and the output in diffusion model. \cite{abstreiter2021diffusion,mittal2023diffusion} propose a time-conditioned encoder and demonstrate the performance on downstream classification tasks. \cite{zhang2022unsupervised} argues that there is a information loss in the forward process when using the pretrained diffusion models for presentation learning and propose to predict the mean shift to fill the loss. 
However, it's important to note that there is no guarantee that the learned representations by references \cite{preechakul2022diffusion,wang2023infodiffusion,zhang2022unsupervised,abstreiter2021diffusion,mittal2023diffusion,hudson2023soda} are the true latent variables. 
Furthermore, unlike VAEs and GANs, diffusion models inherently require numerous iterations to generate data from noise. 
This iterative process could pose challenges in effectively integrating SCMs into diffusion models, potentially impacting their practical use cases.
}



{
In addition to VAEs, GANs and diffusion-based models, normalizing flows also serve as an important element in causal representation learning due to its invertiblility nature.
iFlow \cite{li2019identifying}, rather than optimizing a log-likelihood lower bound like iVAE, employs normalizing flows for direct log-likelihood maximization and recovers the true latents. 
The study by \cite{von2021self} posits data augmentation as a latent model, elucidating the success of contrastive learning by isolating unwanted style information. 
CauCA \cite{liang2023causal} achieves causal mechanism identifiability with intervention datasets and a known causal graph. 
Conversely, \cite{von2023nonparametric} identifies bi-variate latent variables across different datasets without requiring the causal graph, emphasizing the removal of parental influences through perfect interventions. They validate their findings using normalizing flow, selecting the best-performing model.}

%% file: secs/sec_Causality_LLMs.tex
\begin{table*}[!tb]
\caption{A comprehensive categorization of current research on the causal capabilities of LLMs along two dimensions: methodologies and tasks. 
The methodologies refer to the approaches for utilizing LLMs, which can be language comprehension, serving as a knowledge base, or formal reasoning. 
The tasks represent specific causal applications where LLMs are employed.
}
\centering
\begin{tabular}{>{\centering\arraybackslash}p{4cm}>{\centering\arraybackslash}p{4cm}>{\centering\arraybackslash}p{2.6cm}>{\centering\arraybackslash}p{3.2cm}>{\centering\arraybackslash}p{2.3cm}}
\toprule
\textbf{Methodology/Task} & \textbf{Event Causality Identification} & \textbf{Causal Explanation} & \textbf{Causal Discovery} & \textbf{Causal Inference} \\
\midrule
\textbf{LLMs as Language Comprehension} & ~\cite{hobbhahn2022investigating},~\cite{zhang2023understanding},~\cite{pawlowski2023answering},~\cite{kiciman2023causal} & ~\cite{gao2023chatgpt} & -- & --  \\
\midrule
\textbf{LLMs as Knowledge Base} & -- & -- & ~\cite{choi2022lmpriors}, ~\cite{long2023can}, ~\cite{kiciman2023causal}, ~\cite{gao2023chatgpt}, ~\cite{ban2023query}, ~\cite{long2023causal}, ~\cite{willig2023causal} & --\\
\midrule
\textbf{LLMs as Formal Reasoning} & -- & -- & ~\cite{jin2023can} & ~\cite{jin2023cladder} \\
\bottomrule
\end{tabular}

\label{312606125793}
\end{table*}

The advent of large-scale generative models, especially generative large language models (LLMs)
\cite{brown2020language,ouyang2022training,radford2018improving,radford2019language,vaswani2017attention} such as DELL-A, Stable Diffusion models and GPTs \cite{oppenlaender2022creativity,ramesh2022hierarchical,rombach2022high, openai2023gpt4}, have not only markedly advanced the capabilities of conventional DGMs but also redefined the benchmarks for performance and complexity in the field.
This confluence of computational power and nuanced data representation has spurred interest in the causal community. 
Consequently, an emerging line of research is focusing on leveraging LLMs for tasks of causal reasoning. 
This intersection between causality and LLMs constitutes an emergent frontier in artificial intelligence research \cite{kaddour2023challenges}.
In this section, we delve into this rapidly evolving area, posing two pivotal questions. 
First, what exactly are the current causal capabilities of LLMs—do they primarily memorize causal relationships encountered in training data, or do they manifest capacities for genuine causal reasoning ($\S \ref{066666237491}$)? 
Second, how can the architectural complexity and computational scale of LLMs be utilized to advance the methodologies and applications in causal research ($\S \ref{605290963442}$)?
Besides, we explore the unknown of causality in large-scale DGMs for tabular data and time-series (\S \ref{164439834288}).

By examining these questions, we aim to provide a comprehensive overview of the present capabilities and future possibilities of  exploring causality with large-scale generative models.

\subsection{Assessing Causal Capabilities in Current LLMs: Memorization or Reasoning?}
\label{066666237491}
The existing literature on evaluating causal capabilities of LLMs can be stratified along two dimensions:
the methodologies for employing LLMs and the specific causal tasks for which LLMs are utilized, as shown in Table \ref{312606125793}.

We introduce approaches to harness LLMs for causal analysis, drawing inspiration from \cite{jin2023cladder}:
\begin{itemize}
\item \textbf{LLMs as Language Comprehension for Learning Causality:} 
in this paradigm, LLMs are primarily seen as language comprehension systems. 
The focus is on their ability to understand the structure of natural language text and identify causal relationships embedded within it. 
\item \textbf{LLMs as Knowledge Base for Learning Causality:} the emphasis is on the LLM's capability to serve as a repository of knowledge, particularly causal knowledge that can be extracted for discovery purposes. 

\item \textbf{LLMs as Formal Reasoning for Learning Causality:} in this mode, LLMs are expected to engage in formal or semi-formal reasoning to make causal inferences \cite{jin2023cladder}, such as treatment effect estimation \cite{zhou2022cycle,li2022contrastive,zhou2023meta}. 
They would be expected to apply logical or statistical reasoning to existing data to arrive at new causal conclusions. 
\end{itemize}

\subsubsection{LLMs as Language Comprehension}
\begin{figure}[!hb]
  \centering
  \includegraphics[width=0.80\linewidth]{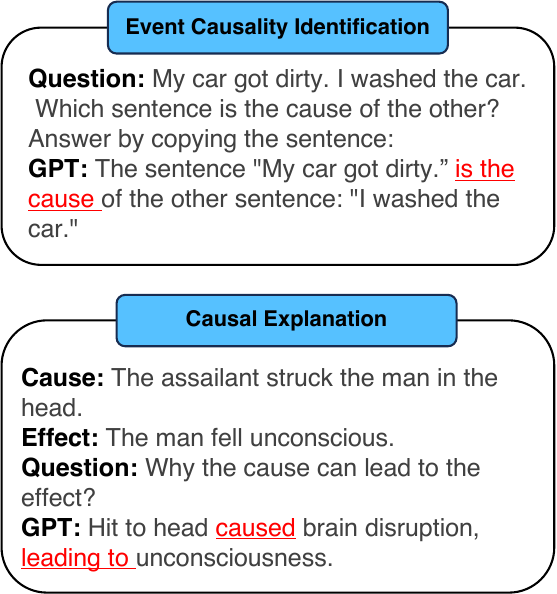}
  \caption{The forms of event causality identification \cite{hobbhahn2022investigating} and causal explanation tasks \cite{gao2023chatgpt}. 
  The content that characterizes the cause is marked in \textbf{
  {\underline{red}}}.}
  \label{034556171337}
\end{figure}

In this paradigm, LLMs are used to identify and explain causal relationships within textual data. 
This methodology excels in tasks like event causality identification and causal explanation generation, leveraging the LLMs' natural language processing capabilities, as shown in Figure \ref{034556171337}. 
It focuses on parsing text to discern explicit or implicit cause-and-effect \cite{zhang2023understanding}.

\begin{table*}[!htb]
\caption{Accuracy results for causal discovery tasks, as reported in \cite{kiciman2023causal,gao2023chatgpt}. Only four widely-referenced datasets are presented.}
\centering
\begin{tabular}{>{\centering\arraybackslash}p{4cm}>{\centering\arraybackslash}p{4cm}>{\centering\arraybackslash}p{4cm}>{\centering\arraybackslash}p{2cm}>{\centering\arraybackslash}p{2cm}}
\toprule
Models & Tubingen cause-effect~\cite{mooij2016distinguishing} & Neuropathic pain~\cite{tu2019neuropathic} & e-CARE~\cite{du2022care} & COPA~\cite{roemmele2011choice} \\
\hline
\textbf{Data-driven} &&&&  \\
PNL-MLP \cite{zhang2012identifiability} & 75.0 & -- & -- & -- \\
Mosaic \cite{wu2020causal} & 83.3 & -- & -- & -- \\ \hline
\textbf{GPT-Based}\footnotemark[8] &&&&  \\
text-davinci-002 & 79.0 & 51.7 & 78.4 & 94.4 \\
text-davinci-003 & 82.0 & 55.1 & 76.7 & 93.2 \\
gpt-3.5-turbo & 81.0 & 71.1 & 79.1 & 96.3 \\
gpt-4 & \textbf{96.0} & \textbf{78.4} & \textbf{84.5} & \textbf{98.1}\\
\bottomrule
\end{tabular}
\label{859484596543}
\end{table*}

\subsubsection{LLMs as Knowledge Base}
In causal reasoning, constructing an accurate causal graph is often the first step \cite{zhang2016discovery,zhang2017causal,huang2017behind,xie2020generalized,zhang2015estimation}. 
Conventional algorithms use conditional independence tests but may need expert inputs, especially in complex domains like healthcare (\S \ref{919611488502}). 
LLMs, with their rich knowledge bases, offer a more efficient alternative.

\begin{center}
\fbox{%
\begin{minipage}{0.97\linewidth}
\textbf{Distinction: "LLMs as Knowledge Base" vs. "LLMs as Language Comprehension"} — Both paradigms involve language comprehension. However, "LLMs as Knowledge Base" goes further by utilizing LLMs' internal knowledge to provide deeper insights into causal relationships that may not be explicit in the query, thereby offering a more nuanced understanding of causality.
\end{minipage}%
}    
\end{center}

A pioneering work in this space \cite{choi2022lmpriors}, 
presents LMPriors, a framework that integrates task-specific priors from LLMs into causal discovery. 
Specifically, the framework utilizes natural language metadata, such as variable names and descriptions, to guide downstream models in generating outputs that are consistent with common-sense reasoning. 
Within the scope of causal discovery, LMPriors aims to identify the directional relationship between two variables, denoted as $\bm{x} \rightarrow \bm{y}$ or $\bm{y} \rightarrow \bm{x}$. 
These LLM-derived priors act as initial steps before applying a traditional, data-driven causal discovery algorithm $f\left(\mathcal{D}\right)$:
\begin{equation}
\label{118902280853}
    \mathcal{P}_{\text{CD}}(f)(\mathcal{D}) = \text{log} \left(\frac{p_{\text{LLMs}}(\bm{x} \rightarrow \bm{y} \vert \bm{c}(\mathcal{D}_{\text{meta}}))}{p_{\text{LLMs}}(\bm{y} \rightarrow \bm{x} \vert \bm{c}(\mathcal{D}_{\text{meta}}))} \right) + f(\mathcal{D})
\end{equation}
where $\bm{c}(\mathcal{D}_{\text{meta}})$ serves as a prompt designed to elicit a directional relationship, and $\mathcal{P}_{\text{CD}}(f)$ returns the posterior probability of the most likely causal structure between $\bm{x}$ and $\bm{y}$.

Concurrent studies \cite{long2023can,ban2023query,long2023causal,kiciman2023causal} underscore the potential of LLMs, particularly GPT-3.5 and 4, in advancing causal research.
We showcase several such evaluations in Table \ref{859484596543}. 
These works integrate LLMs' expertise with data-driven methods, achieving state-of-the-art performance on various causal benchmarks and proposing new LLMs-based causal analysis pipelines, as shown in Figure \ref{109097681712}.
Contrary to previous assessments, \cite{gao2023chatgpt,willig2023causal} find ChatGPT performs poorly across three causal reasoning tasks. 
They introduce a \textit{binary classification} setting alongside \textit{multiple choice} for causal discovery. Their results indicate that ChatGPT\footnote{https://chat.openai.com/} excels at identifying causal pairs but falters in recognizing non-causal pairs. This challenges prior work \cite{kiciman2023causal}, which used only multiple-choice tests and thereby overestimated ChatGPT's causal reasoning abilities.

\subsubsection{LLMs as Formal Reasoning}
Diverging from the realms of empirical knowledge and natural language comprehension, \cite{jin2023can, jin2023cladder} focus on the capacity of LLMs for formal causal reasoning, an essential aspect of human cognitive processes. 
In contrast to extracting or interpreting pre-existing causal knowledge, formal causal reasoning entails generating logically sound causal inferences under varying conditions. 
To scrutinize LLMs' prowess in this specialized domain, the authors introduce two benchmark datasets, corr2cause\footnotemark[9] and cladder\footnotemark[10], designed to assess a wide range of causal reasoning skills, from basic associative understanding to advanced counterfactual analysis. 
They also present CAUSALCOT, a novel chain-of-thought prompting strategy, which markedly improves LLM performance on these benchmarks. {
A related work \cite{tang2023towards} proposes CaCo-CoT, a chain-of-thought method aimed at enhancing knowledge-based reasoning in LLMs by improving causal consistency. Furthermore, another study  \cite{ji2023benchmarking} focuses on enhancing the quality of code generated by LLMs. They achieve this by first constructing a causal graph to identify the causal relationships between prompts and the generated code and subsequently calibrating the input prompts effectively.}
Despite these advancements, the study underscores the limitations of LLMs in formal causal reasoning and sets the stage for future work aimed at enhancing their causal capabilities.

\subsubsection{Memorization or Reasoning?}

A prevailing challenge in evaluating LLMs on causal tasks is the issue of data contamination. 
This phenomenon arises when LLMs excel on a test set due to inadvertent inclusion of test data in the training set. 
To distinguish between memorization and true reasoning capabilities, two notable approaches have been adopted:

\begin{itemize}
  \item \textbf{Memorization Tests:} Kiciman et al. \cite{kiciman2023causal} implement a memorization test where LLMs are supplied with initial columns from a dataset, including row ID and variable names, and are prompted to complete the remaining columns. 
  Such test for Tubingen cause-effect dataset~\cite{mooij2016distinguishing} reveals that GPT-3.5 can accurately recall 58\% of the remaining cells and 19\% of entire rows without error. 
  GPT-4 performs marginally better, recalling 61\% of cells and 25\% of entire rows. 
  These results suggest that although the dataset is likely part of GPT's training data, there remains a significant gap between memorization and overall accuracy.
  \item \textbf{Isolation of Data Contamination Effects:} Jin et al. \cite{jin2023cladder} develop a dataset with variations designed to isolate memorization effects. 
  They employ a verbalization procedure that transforms symbolic variables into natural language narratives to describe causal processes. 
  To mitigate data contamination, they create commonsensical, anti-commonsensical, and nonsensical versions of the dataset. 
  GPT-4 performs best on the commonsensical data but shows a 5.34-point decline on the nonsensical version, indicating that memorization contributes minimally to its performance.
\end{itemize}

These approaches aim to offer more robust assessments of an LLMs' true causal reasoning capabilities by mitigating the influence of data contamination. 
While the findings confirm that LLMs do exhibit memorization, this memorization has a marginal influence on their reasoning performance. 
This suggests a promising indication of LLMs' reasoning capabilities. 
However, for a precise estimation, further comprehensive assessments are necessary.

\begin{figure}[!b]
  \centering
  \includegraphics[width=0.97\linewidth]{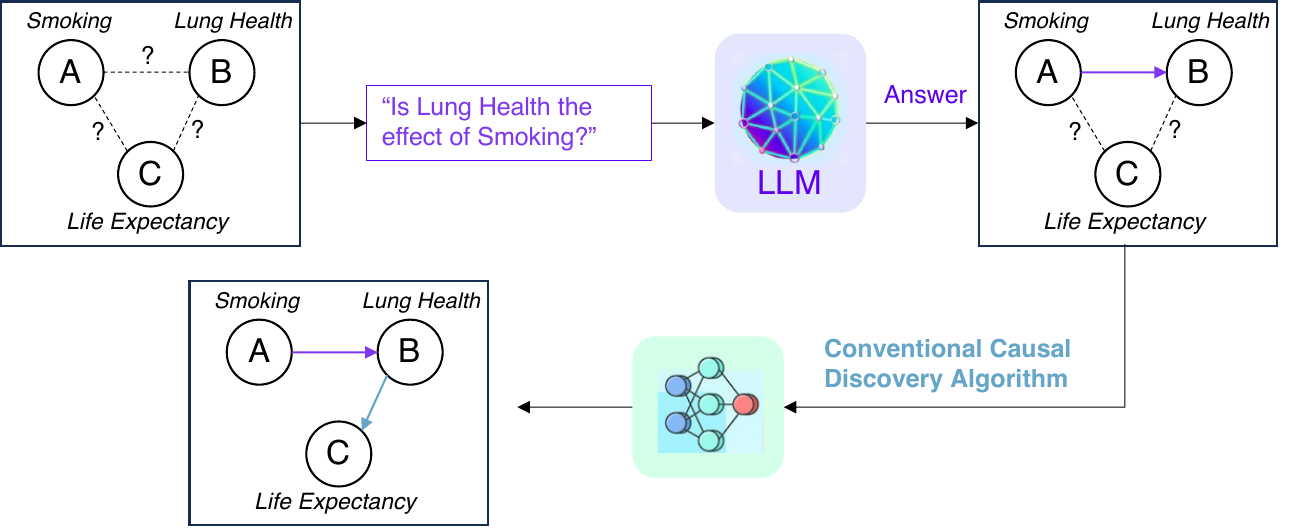}
  \caption{Illustration of employing LLMs as a knowledge repository, complemented by data-driven algorithms for causal discovery \cite{willig2023causal,bommasani2021opportunities}. 
  This establishes a LLMs-based causal analysis pipeline \cite{kiciman2023causal}.}
  \label{109097681712}
\end{figure}

\subsection{Advancing Causal Research Through LLMs}
\label{605290963442}

\subsubsection{Combining LLMs with Data-Driven Methods}
\label{360858045108} 
The inherent capabilities of LLMs to identify complex patterns in extensive datasets position them as a valuable complement to existing algorithmic methods in causal discovery and inference \cite{kiciman2023causal}. 
Serving initially as a pre-processing mechanism, LLMs can analyze observational data to identify potential causal relationships. 
These preliminary findings can subsequently be subjected to rigorous validation and quantification using conventioal data-driven causal algorithms, as illustrated in Figure \ref{109097681712}. 
Beyond this initial phase, the LLMs can be further integrated into the causal analysis pipeline, serving to refine and enhance the outputs generated by conventional causal algorithms.

\subsubsection{Crafting Effective Prompts for Causal Reasoning}
\label{688524414352}
Effective prompts are able to guide the generation of more accurate and interpretable outputs in LLMs \cite{shin2020autoprompt, wei2022chain}.
In addition, well-crafted prompts also serve as a mechanism for querying the LLMs' understandings of causal relationships. 
A well-designed prompt may help in extracting the LLMs' implicit knowledge about confounders, mediators, or effects in a given causal pathway, thereby aiding the interpretation and validation of empirical causal models.
The crucial role of effective prompting has been underscored in multiple studies \cite{long2023can,gao2023chatgpt,jin2023cladder}.

\subsection{Exploring the Unknown: Causality in Large-Scale DGMs for Tabular Data and Time-Series}
\label{164439834288}
{
The exploration of causality within large-scale DGMs has been prominently focused on language models, leaving the potential in tabular data and time-series relatively untapped. We recognize this as a significant gap in the current research landscape. Causality offers a framework for generating more representative and balanced datasets, particularly vital in scenarios with missing or skewed data, as seen in studies like \cite{chen2019faketables, camino2020working, engelmann2021conditional}. This approach could be transformative in areas like healthcare and finance, where decision-making often depends on complex tabular datasets~\cite{borisov2022deep} and time-series data~\cite{yao2022temporally, yao2021learning}, highlighting a promising avenue for future research.
}